\documentclass[twoside]{article}
\pdfoutput=1
\usepackage[accepted]{aistats2023}

\usepackage[utf8]{inputenc}
\usepackage{amsmath}
\usepackage{amssymb}
\usepackage{mathtools}
\numberwithin{equation}{section}
\usepackage{slashed}
\usepackage{braket}
\usepackage{enumitem}
\usepackage[bottom]{footmisc}
\usepackage{cite}
\usepackage{graphicx}
\usepackage{bm}
\usepackage{subfig}
\usepackage{physics}
\usepackage{xcolor}
\usepackage{natbib}
\usepackage{mdframed}

\usepackage{CJK}

\usepackage[linesnumbered,ruled]{algorithm2e}
\SetKwComment{Comment}{/* }{ */}
\SetKwInput{kwInit}{Initialization}
\SetKwRepeat{Do}{do}{while}%

\newenvironment{claim}{  \begin{mdframed}[linecolor=black!0,backgroundcolor=black!10]\noindent%
		\ignorespaces}{\end{mdframed}}

\renewenvironment{figure}[1][]{
  \begin{originalfigure}[#1]
    \begin{mdframed}[linecolor=black!0,backgroundcolor=black!1]
}{
    \end{mdframed}
  \end{originalfigure}
}

\def\s{\sigma }

\def\0{{(0)}}
\def\1{{(1)}}
\def\2{{(2)}}

\def\<{\langle }
\def\>{\rangle }

\newcommand{\bea}{\begin{eqnarray}}
\newcommand{\eea}{\end{eqnarray}}

\usepackage{slashed}

\def\({\left(}
\def\){\right)}
\def\[{\left[}
\def\]{\right]}

\usepackage{dashbox}
\usepackage{xcolor}
\usepackage{colortbl}
\usepackage{arydshln}
\definecolor{lightyellow}{rgb}{1.0, 0.95, 0.7}
\definecolor{Blue}{rgb}{0.54, 0.17, 0.89}
\definecolor{blue-violet}{rgb}{0.54, 0.17, 0.89}
\definecolor{blue}{rgb}{0,0,1}
\definecolor{darkgreen}{rgb}{0.,0.6,0.}
\definecolor{green}{rgb}{0.0, 0.42, 0.24}

\newcommand*{\Blue}[1]{\textcolor{Blue}{#1}}
\newcommand*{\blue}[1]{\textcolor{blue}{#1}}
\newcommand*{\green}[1]{\textcolor{green}{#1}}

\definecolor{colorA}{rgb}{1,0,0}
\definecolor{colorB}{rgb}{0,0.3,1}
\definecolor{colorC}{rgb}{0.9,0.8,0.2}
\definecolor{colorD}{rgb}{0,0.65,0}
\definecolor{lesslightgray}{rgb}{0.5,0.5,0.5}

\definecolor{light-gray}{gray}{0.95}
\newcommand{\code}[1]{\colorbox{light-gray}{\footnotesize\texttt{#1}}}

\let\tilde\widetilde
\let\hat\widehat

\newcommand{\calE}{\mathcal{E}}
\newcommand{\calF}{\mathcal{F}}
\newcommand{\calG}{\mathcal{G}}

\newcommand{\calO}{\mathcal{O}}

\newcommand{\calR}{\mathcal{R}}
\newcommand{\calS}{\mathcal{S}}

\newcommand{\calU}{\mathcal{U}}
\newcommand{\calV}{\mathcal{V}}

\newcommand{\calX}{\mathcal{X}}

\newcommand{\calZ}{\mathcal{Z}}

\newcommand{\bO}{\bm{O}}

\newcommand{\bX}{\bm{X}}
\newcommand{\bY}{\bm{Y}}

\newcommand{\bx}{\bm{x}}
\newcommand{\by}{\bm{y}}

\newcommand{\Max}{\mathop{ \rm Max}}
\newcommand{\Min}{\mathop{\rm Min}}

\newcommand{\E}{\mathop{\mathbb{E}}}

\let\cite\citep 

\newcommand{\sumN}{\sum_{i=1}^N}

\newcommand{\sumX}{\sum_{\{\bX\}}}

\usepackage{amsthm}
\usepackage{nicefrac}

\newtheorem{theorem}{Theorem}[section]
\newtheorem{corollary}{Corollary}[section]
\newtheorem{lemma}[theorem]{Lemma}
\newtheorem{remark}{Remark}[section]
\theoremstyle{definition}
\newtheorem{definition}{Definition}[section]

\usepackage{amssymb,amsmath,amsthm,enumitem}

\numberwithin{equation}{section}
\numberwithin{theorem}{section}

\usepackage{lipsum}

\newcommand*{\annot}[1]{\tag*{\footnotesize\textcolor{black!50}{#1}}}

\makeatletter
\let\save@mathaccent\mathaccent
\newcommand*\if@single[3]{%
    \setbox0\hbox{${\mathaccent"0362{#1}}^H$}%
    \setbox2\hbox{${\mathaccent"0362{\kern0pt#1}}^H$}%
    \ifdim\ht0=\ht2 #3\else #2\fi
}
\newcommand*\rel@kern[1]{\kern#1\dimexpr\macc@kerna}
\newcommand*\widebar[1]{\@ifnextchar^{{\wide@bar{#1}{0}}}{\wide@bar{#1}{1}}}
\newcommand*\wide@bar[2]{\if@single{#1}{\wide@bar@{#1}{#2}{1}}{\wide@bar@{#1}{#2}{2}}}
\newcommand*\wide@bar@[3]{%
    \begingroup
    \def\mathaccent##1##2{%
        \let\mathaccent\save@mathaccent
        \if#32 \let\macc@nucleus\first@char \fi
        \setbox\z@\hbox{$\macc@style{\macc@nucleus}_{}$}%
        \setbox\tw@\hbox{$\macc@style{\macc@nucleus}{}_{}$}%
        \dimen@\wd\tw@
        \advance\dimen@-\wd\z@
        \divide\dimen@ 3
        \@tempdima\wd\tw@
        \advance\@tempdima-\scriptspace
        \divide\@tempdima 10
        \advance\dimen@-\@tempdima
        \ifdim\dimen@>\z@ \dimen@0pt\fi
        \rel@kern{0.6}\kern-\dimen@
        \if#31
        \overline{\rel@kern{-0.6}\kern\dimen@\macc@nucleus\rel@kern{0.4}\kern\dimen@}%
        \advance\dimen@0.4\dimexpr\macc@kerna
        \let\final@kern#2%
        \ifdim\dimen@<\z@ \let\final@kern1\fi
        \if\final@kern1 \kern-\dimen@\fi
        \else
        \overline{\rel@kern{-0.6}\kern\dimen@#1}%
        \fi
    }%
    \macc@depth\@ne
    \let\math@bgroup\@empty \let\math@egroup\macc@set@skewchar
    \mathsurround\z@ \frozen@everymath{\mathgroup\macc@group\relax}%
    \macc@set@skewchar\relax
    \let\mathaccentV\macc@nested@a
    \if#31
    \macc@nested@a\relax111{#1}%
    \else
    \def\gobble@till@marker##1\endmarker{}%
    \futurelet\first@char\gobble@till@marker#1\endmarker
    \ifcat\noexpand\first@char A\else
    \def\first@char{}%
    \fi
    \macc@nested@a\relax111{\first@char}%
    \fi
    \endgroup
    }
\makeatother
\let\bar\widebar

\definecolor{DarkGreen}{rgb}{0,0.40,0}
\definecolor{FireBrick}{rgb}{0.698,0.133,0.133}
\usepackage[colorlinks,citecolor=DarkGreen,linkcolor=FireBrick,urlcolor=FireBrick]{hyperref}
\usepackage{url}
\usepackage{booktabs}
\usepackage{lipsum}
\usepackage{titlesec}
\usepackage{wrapfig,lipsum,booktabs}
\titlespacing\section{0pt}{8pt plus 4pt minus 2pt}{0pt plus 2pt minus 2pt}
\titlespacing\subsection{0pt}{6pt plus 4pt minus 2pt}{0pt plus 2pt minus 2pt}
\titlespacing\subsubsection{0pt}{6pt plus 4pt minus 2pt}{0pt plus 2pt minus 2pt}

\begin{document}

\runningtitle{\textsc{CoRMF}: Criticality-Ordered Recurrent Mean Field Ising Solver}

\runningauthor{
    Zhenyu Pan,
    Ammar Gilani,
    En-Jui Kuo,
    Zhuo Liu
    }

\twocolumn[

\aistatstitle{\textsc{CoRMF}: Criticality-Ordered Recurrent Mean Field Ising Solver
}

\aistatsauthor{  
         Zhenyu Pan$^{1}$,
         Ammar Gilani$^{2}$,
         En-Jui Kuo$^{3,}$,
         Zhuo Liu$^{4}$
        }

\aistatsaddress{
        $^{1,4}$University of Rochester;
        $^{2}$Northwestern University;
        $^{3}$University of Maryland, College Park; \\
        $^{3}$National Center for Theoretical Sciences (Physics Division); $^{3}$HonHai (Foxconn) Research Institute, Taipei, Taiwan\\
        \texttt{\footnotesize
        \texttt{\small\{$^{1}$\href{mailto:zhenyupan@rochester.edu}{zhenyupan},$^{4}$\href{mailto:zhenyupan@rochester.edu}{zhuo.liu}\}@rochester.edu};
        \{$^{2}$\href{mailto:Ammargilani2024@u.northwestern.edu}{ammar}\}@u.northwestern.edu;}\\
        \texttt{\{$^{3}$\href{mailto:kuoenjui@umd.edu}{kuoenjui}\}@umd.edu}}
        ]
        
\begin{abstract}
We propose an RNN-based efficient Ising model solver, the \textbf{C}riticality-\textbf{o}rdered \textbf{R}ecurrent \textbf{M}ean \textbf{F}ield (CoRMF), for forward Ising problems.
In its core, a criticality-ordered spin sequence of an $N$-spin Ising model is introduced by sorting mission-critical edges with greedy algorithm, such that an autoregressive mean-field factorization can be utilized and optimized with Recurrent Neural Networks (RNNs).
Our method has two notable characteristics: (i) by leveraging the approximated tree structure of the underlying Ising graph, the newly-obtained criticality order enables the unification between variational mean-field and RNN, allowing the generally intractable Ising model to be efficiently probed with probabilistic inference; (ii) it is well-modulized, model-independent while at the same time expressive enough, and hence fully applicable to any forward Ising inference problems with minimal effort. Computationally, by using a variance-reduced Monte Carlo gradient estimator, CoRFM solves the Ising problems  in a self-train fashion without data/evidence, and the inference tasks can be executed by directly sampling from RNN. Theoretically, we establish a provably tighter error bound than naive mean-field by using the matrix cut decomposition machineries. Numerically, we demonstrate the utility of this framework on a series of Ising datasets.

\end{abstract}

\setlength{\abovedisplayskip}{4pt}
\setlength{\belowdisplayskip}{4pt}
\setlength{\parskip}{0.3em}
\vspace{-2mm}
\section{Introduction}

\label{sec:intro}
The exact computation of Ising models are known to be NP problems for classical computers in CS community \cite{mezard2009information,barahona1982computational}, and they are connected to all other NP (-complete and -hard) problems through some easy-to-generalize methods \cite{de2016simple,lucas2014ising}.

On one hand, the connection between NP problems and Ising models has resulted in strong physics intuitions \cite{kirkpatrick1983optimization} that the hardness of these problems emerges through the lens of complex energy landscapes over discrete random variables with multiple local minima \cite{chowdhury2014spin,barahona1982computational}.
On the other hand, the computational difficulty on the Ising side resonates with the difficulties of numerous significant scientific problems, including numerous other combinatorial decision-making and optimization problems \cite{benati2007mixed,ngo1994computational,garey1979computers}.
As the opposite of conventional inverse Ising problems \cite{nguyen2017inverse,reneau2023feature} that reconstruct graphical structure from data, we refer to these problems, which have pre-specified graphical structures, as \textit{forward Ising problems} (combinatorial inference and optimization problems in Ising formulations \cite{de2016simple,lucas2014ising,pan2023ising}), and any efficient computational method or hardware solver \cite{mohseni2022ising} for Ising models can potentially benefit them. 

To describe the Ising model, we first introduce some notation here.
We consider an Ising model of $N$ spins as an exponential family model for binary $N$-spin data up to quadratic sufficient statistic taking the Boltzmann form
\footnotesize
\begin{align}
    \label{eqn:boltz}
    P(\bX) 
    &= \frac{1}{\calZ} \exp\left\{-\beta E(\bX)\right\}\\
    &= \frac{1}{\calZ}\exp\left\{-\beta \(\sum_{e_{ij}\in \calE} J_{ij}\bx_i\bx_j + \sum_{v_i\in \calV} h_i \bx_i\)\right\}
    \,,\nonumber
\end{align}\normalsize
where $\bX\coloneqq\{\bx_1,\cdots,\bx_N\}\in\{\pm1\}^N$ is the configuration of $N$ binary random variables (spins) $\bx_i\in\{\pm1\}$ assigned to the Ising graph $\calG = (\calV,\calE)$ , $\beta\ge 0$ is the inverse temperature, and $\calZ\coloneqq \sum_{\bX} e^{-\beta E(\bX)}$ is the partition function ensuring the normalization of $P(\bX)$.
The graphical structure of the Ising model (and corresponding Ising problem) is encoded into the Ising energy function $E(\bX)$ through the symmetric $N\times N$ strength matrix $J$ with zeros on the diagonal, and the external field vector $h$.

To get a taste for the forward Ising problem, we consider the following NP problem as an example, the Number Partitioning Problem (NPP), and its Ising formulation.
\begin{claim}\vspace{-2mm}
    \begin{definition}[Number Partitioning Problem]
    Given a set of $N$ positive numbers $\calS=\{n_i\}_{i=1}^N$, ask whether there is a partition of $\calS$ into two disjoint sets $\calR$ and $\calS - \calR$, such that the sum of the elements in both sets is equal.
    What is the partition if it exists?
\end{definition}
\end{claim}
The problem is clearly NP, and its corresponding Ising model can be identified as 
$E(\bX) = \(\sumN n_i \bx_i\)^2$ whose $(J,h)$ can be obtained easily.
If there exists a configuration $\bX^\star$ such that $E\(\bX^\star\)=0$, $\bX^\star$ provides the desired parition.
Moreover, $E = 0$ is the ground state of the Ising model (the state with lowest energy) and therefore finding the solution to NPP is equivalent to solving the ground state of the Ising model.

In this work, rather than concentrating on the Ising formulation side, we are interested in solving a subclass of the forward problems, the \textit{forward Ising inference}, 
with variational mean-field method.
In particular, we focus on solving the variational mean-field inference problems of pre-specified Ising models $(J,h)$ (presumably transformed from NP problems) in the setting where no observed data is available.
Using above NNP example, with such an energy function provided, we intend to do predictive inference on it through the intractable \eqref{eqn:boltz} with variational mean-field method, i.e. inferring marginal, conditional or mode information of the NP problem.

Our goal is to develop a variational mean-field Ising solver, a method that can be efficiently applied to any forward Ising inference problems with minimal effort.
That is, we aim to approximate the notoriously intractable \textit{target} distribution $P(\bX)$ \cite{sly2012computational,istrail2000statistical,barahona1982computational} with a variational mean-field \textit{model} distribution $Q_\theta$ by minimizing the Kullback-Leibler (KL) divergence,
such that inference tasks can be solved tractably and efficiently.
To this end,
we propose a new type of variational mean-field integrated with recurrent neural networks (RNNs), the Criticality-ordered Recurrent Mean-Field (CoRMF).
CoRMF generalizes the standard mean field approach by employing the \textit{criticality-ordered autoregressive factorization} for the modeled distribution $Q_\theta$ 
\bea
Q_{\theta}(\bX) = \prod_{i=1}^N q_{\theta}\(\bx_i|\bx_{i-1},\cdots,\bx_{1}\)
=\prod_{i=1}^N q_{\theta}\(\bx_i|\bX_{j<i}\),\label{AR_ansatz}
\eea
and parametrize it with an RNN, 
where the criticality-ordered spin sequence, $\bX_{j<i}=(\bx_1,\cdots,\bx_N)$, of an $N$-spin Ising model is ordered by 
sorting mission-critical edges using greedy algorithm.
That is, the edges are ordered according to their criticality by exploiting the tree approximation to the Ising graph, and the autoregressive decomposition \eqref{AR_ansatz} is constructed based on this relative order throughout this paper.

As for optimization, in contrast to traditional variational Bayes mean-field methods \cite{wainwright2008graphical,attias1999variational} that optimize $Q_\theta$ iteratively with some observed data,
we perform the optimization in a self-train fashion by leveraging the direct sampling capability of RNN \cite{graves2013generating,bengio2017deep} combining with a variance-reduced Monte Carlo gradient estimator to circumvent the lack of training data and the intractability to backpropagate through samples of discrete random variables with auto-differentiation.
Moreover, as for inference, the learned mean-field $Q_\theta$ may be used to directly generate unbiased uncorrelated samples and their corresponding probabilities (neural samplers \cite{nicoli2020asymptotically,wu2019solving,nowozin2016f,goodfellow2016deep}).

Theoretically, we demonstrate that the negative-log-partition or the variational free energy of CoRMF is restricted by tighter error bound than Naive Mean-Field (NMV) for general Ising graphs by using the matrix cut decomposition machineries \cite{jain2018mean,frieze1999quick}.

\paragraph{Contributions.}
\vspace{-2mm}
\begin{itemize}\vspace{-2mm}
\setlength\itemsep{0em}
    \item 
    Our method serves as a generic variational mean-field framework for solving forward Ising problems by combining the power of information extraction at the graph level (criticality-ordered autoregressive decomposition) and sequential learning at the architecture level (RNN.)
    
    \item 
    The CoRMF  significantly reduces the number of parameters  required to parametrize all the conditionals from exponential number to polynomial (to the size $N$), while at the same time the expressiveness is supported by the universal approximation theorem.
    
    \item
    We demonstrate a provably tighter error bound than the naive mean-field by using the matrix cut decomposition machineries.
    
    \item 
    We numerically validate CoRMF on 4 sets of Ising model (with given $(J,h)$).
    We demonstrate that our method obtained better error bounds than naive mean-field and CoRMF with \textit{random \& reverse} criticality-ordered spin sequences. 
    \vspace{-2mm}
\end{itemize}

The paper is organized as follows. 
In Sec.~\ref{sec:method} we present CoRMF, an RNN-based variational mean-field method.
In Sec.~\ref{sec:theory}, the mean-field approximation analysis is extended to our CoRMF.
In Sec.~\ref{sec:exp}, numerical studies are conducted.
Finally, concluding discussions are given in Sec.~\ref{sec:conclusion}.

\vspace{-2mm}
\section{Backgrounds}
\label{sec:background}

When solving statistical physics systems like forward Ising problems, we are usually interested in the macroscopic properties of the system e.g. the ground state energy (the mode), magnetizations (the statistics) or the free energy $\calF \coloneqq - \ln \calZ/\beta$, which could be all deduced from the Boltzmann distribution \eqref{eqn:boltz}.
Equivalently, by solving, we typically refer to combinatorial optimization or inference.
The focus of this work is the latter, namely  forward Ising inference.

\paragraph{Variational Mean-Field Methods.}
\label{sec:VMF}

From the computational perspective, 
inferring \eqref{eqn:boltz} is highly non-trivial.
It is well-established that the exact computation of the Ising partition function $\calZ$ (or equivalently the free energy $\calF$) is NP-hard \cite{mezard2009information, istrail2000statistical,barahona1982computational}, and the approximate sampling and approximating the partition function with high probability and arbitrary precision is also NP-hard \cite{sly2012computational,chandrasekaran2012complexity} for general Ising graphs.

To circumvent the computational hardness, variational mean-field (VMF) provides a way to perform efficient inference by transforming the inference problem into an optimization problem \cite{wainwright2008graphical,attias1999variational}.
In VMF, we first \textit{picks} a $\theta$-parametrized mean-field model family $Q_{\theta}(\bX)$s.
Ideally, the mean-field model $Q_\theta$’s should be expressive enough to resembles the Boltzmann' distribution $P(\bX)$, easy to optimize, and allow for tractable inference the statistical quantities of our interest.
Then we  minimize the KL-divergence
\footnotesize
\bea 
\label{eqn:KL}
 \textsf{KL}\(Q_{\theta}(\bX)|P\)
=\sum_{\bX}Q_{\theta}(\bX) \ln\frac{Q_{\theta}(\bX)}{P(\bX)}
=\beta\(\calF_{\theta}-\calF\),
\eea \normalsize
which is equivalent to minimizing the \textit{variational free energy}, 
\footnotesize
\begin{align}
    \calF_{\theta}
    &\coloneqq\frac{1}{\beta}\sum_{\{\bX\}}Q_{\theta}(\bX)\[\beta E(\bX)+\ln Q_{\theta}(\bX)\]
    \nonumber\\
    &=
    \E_{\bX\sim Q_\theta}\[E(\bX) + \frac{1}{\beta}\ln Q_{\theta}(\bX)\],\label{eqn:free_energy}
\end{align}\normalsize
where $E(\bX)$ is the Ising energy function and $\calF \coloneqq -1/\beta \ln \calZ$ is the \textit{true} free energy\footnote{Note that $\calF_{\theta}$ gives an upper bound to $\calF$.
This work follows the notations in physics community, and hence differs from the traditional Evidence Lower Bound (ELBO) literature with a minus sign.
}.
Throughout this paper, we denote the resulting optimal parameters as $\theta^\star$, its corresponding model $Q_{\theta^\star}$ and minimized variational free energy $\calF_\theta^\star$.

Nevertheless, standard gradient-based optimization method is inapplicable to \eqref{eqn:free_energy} due to the discrete nature of Ising models.
This is mainly because the inability of backpropagating discrete samples or equivalently computing the gradient of \eqref{eqn:free_energy}.
To handle this, relaxations to continuous domain \cite{tucker2017rebar,maddison2016concrete,han2020stein} or reinforce-type gradient estimators \cite{mnih2016variational,ranganath2013black} have been proposed.

\vspace{-2mm}
\section{Methodology}

\label{sec:method}

In this section, we explore the integration of the autoregressive factorization \eqref{AR_ansatz} with RNNs to perform variational mean-field inference.
We first propose a novel relative order among Ising spins in Sec.~\ref{sec:critical_order}, which is then employed to parameterize the variational free energy in \eqref{eqn:free_energy} with an RNN in Sec.~\ref{sec:CoRMF}.
Lastly we discuss the optimization and sampling of the variational problem using a variance-reduced gradient estimator in Sec.~\ref{sec:training}.

\subsection{Criticality-Ordered Spin Sequence}
\label{sec:critical_order}

The spin sequence $\bX_{j<i}=(\bx_{i-1},\cdots,\bx_{1})$ in \eqref{AR_ansatz} indicates a \textit{pre-determined order} for sending the \textit{unordered} spins configuration $\bX=\{\bx_i\}_{i=1}^N$ into the RNN sequentially.
However, since Ising models are inherently permutation-invariant among spins, 
any arbitrary spin ordering would likely compromise the use of \eqref{AR_ansatz} as a mean-field parametrization.

To address this challenge, we present a 
meaningful relative spin order for any given Ising model obtained by greedily searching for mission-critical edges (from high importance to low importance according to $J$) with a modified (and negated) Kruskal's algorithm \cite{graham1985history}.
Beginning with sorting the edges of a given Ising graph into non-increasing order by a heuristic weight $|J_{ij}|$, we let $B$ as the set of edges comprising the tree with the maximal weight spanning and $A$ as the relative order of nodes associated with $B$.
Then, $A$ and $B$ are initialized to be empty, and we start adding edges to $B$ greedily while recording the node order in $A$ until the output sequence is obtained. 

We call the produced sequence as the ``\textit{criticality-ordered spin sequence}." 
In particular, we emphasize that  it is ordered according to the criticality of the edges by  exploiting the tree approximation of the Ising graph, and the autoregressive decomposition \eqref{AR_ansatz} is constructed based on this order throughout this paper.
We refer the pseudocode and details of this greedy algorithm to Appendix~\ref{sec:greedy_algo}.

\subsection{Criticality-Ordered Recurrent Mean-Field} 
\label{sec:CoRMF}

With the criticality-ordered mean-field \eqref{AR_ansatz}, we parametrize $Q_\theta$ with an RNN by treating the
spin values as binary random variables using a softmax layer in the conditional distributions.
Therefore, each node's realization $\hat{\bx}_i$ in the output sequence is associated with the $i$'th conditional 
\bea
q_\theta(\bx_i=\hat{\bx}_i|\bX_{j<i})
=\mathsf{Softmax}\(\bO_i\),
\label{softmax}
\eea
where $\bO_i$ is the $i$th output layer according to the criticality-order spin sequence.
Since the softmax function ranges between $(0,1)$, for each input configuration $\bX$, the joint distribution $Q_\theta(\bX)$ can be decomposed in this way that the each output  conditional is the probability of $\bx_i$ being $\hat{\bx}_i=\pm1$, given the conditioned configuration of spins in front of it ($\bX_{j<i}$).\footnote{Note that the choice of first spin realization ${\hat{\bx}_1}$ is arbitrary due to the parity symmetry of Ising spins.}
With this parametrization, the mean-field \eqref{eqn:free_energy} can be optimized by simply
training the RNN.

Moreover, \eqref{AR_ansatz} and \eqref{softmax} together can be understood as a generalized mean-field, the Criticality-ordered Recurrent Mean-Field (CoRMF), with following features:
(i) it is general to all Ising models and beyond, as the mean-field values in the original Ising quadratic terms (and possible extensions to higher-order terms in more general discrete graphical models) are encoded in RNN weights;
and (ii) the total number of parameters are significantly reduced from exponential in $N$ to polynomial: $(\text{\# of RNN parameters})\propto N^2$; 
(iii) the expressiveness of the mean-field \eqref{AR_ansatz} is guaranteed by the universal approximation theorem;
and (iv)
it provides an efficient surrogate model
for NP problems whose global solution can only be exhaustively solved in exponential time, as the NN is guaranteed to attend local minimum in polynomial time.
Consequently, the CoRMF $Q_\theta$  is fairly expressive, optimizable and applicable for tractable inference.

\subsection{Optimizing CoRMF with RNN Sampler}
\label{sec:training}

Having an optimizable $Q_\theta$ at hand, our next step is to vary the mean-field by training the RNN and minimizing the variational free energy \eqref{eqn:free_energy}.
Here we remind the readers that gradient-based optimization methods are inapplicable to Ising problems and, differed from majority of existing Bayes methods \cite{wainwright2008graphical} that use mean-field to model \textit{conditionals} $q(\bx | \by)$ over random variables $\bx$ with evidence $\by$, the problem of our interest does not involve any observed data.
They make the training (or optimization) more challenging.

To combat these issues, we utilize the reinforce-style gradient estimating algorithm  \cite{mnih2016variational} 
and RNN's ability to draw I.I.D samples directly from $Q_\theta$ \cite{bengio2017deep,graves2013generating} to optimize \eqref{eqn:free_energy} in a self-training manner.
Namely, the RNN is updated with RL-style gradient estimators which are computed with samples drawn from the RNN.

\vspace{-2mm}
\paragraph{RNN Sampler.} 
Due to the design nature of RNN, sampling I.I.D. samples is straightforward.
Since we have the access to all approximated joint conditional probabilities of a given RNN, the sampling directly follows the factorization of the conditional probabilities \eqref{AR_ansatz}, in a predetermined order (the criticality order in this paper) from first to last.
Specifically, to sample a particular configuration $\hat{\bX}= (\bx_1=\hat{\bx}_1,\bx_2=\hat{\bx}_2,\cdots,\bx_N=\hat{\bx}_N)$ where $\hat{\bx}_i$'s are realizations of $\bx_i$'s, we first fixed $\bx_1$ by sampling from its conditional $q(\bx_1)$, then sampling $\bx_2$ from its conditional $q(\bx_2|\hat{\bx}_1)$.
Iteratively repeating the same procedure, we will obtain the sampled configurations $\hat{\bX}$'s (and their corresponding joint $Q_\theta$ and conditional probabilities.)
Note that, the RNN sampler does not have to be trained, and the samples' distribution will progressively tend to $P$ during training.
As we shall see next, RNN sampler enables us to perform self-training with gradient estimator, and generate massive I.I.D. samples for tractable inference after training.

\paragraph{Reinforce Gradient Estimator.}
To train the network, we introduce the Monte Carlo gradient estimator \cite{ranganath2013black} of batch size $K$, 
\footnotesize
\begin{align}
    \beta \nabla_{\theta} \mathcal{F}_\theta 
    &= \nabla_{\theta} \sum_{\bX}  Q_\theta (\bX) \cdot \[ \beta E(\bX) + \ln Q_\theta (\bX)\]
    \nonumber\\
    &=
    \E_{\bX \sim Q(\bX)} \left[ \left( \beta E(\bX) + \ln Q_\theta (\bX) \right) \cdot \nabla_{\theta} \ln Q_\theta (\bX)  \right] 
    \nonumber\\
    &\simeq
    \frac{1}{K}\sum_{k=1}^K \[ \beta E(\bX_k) + \ln Q_\theta (\bX_k) \] \cdot \nabla_{\theta} \ln Q_\theta (\bX_k)  
    ,\nonumber
\end{align}\normalsize
where the contribution of each sample $\bX \sim Q(\bX)$ to the gradient of $\calF_\theta$ is equal to the score function $\nabla_{\theta} \ln Q_\theta (\bX)$ weighted by the reward function\footnote{
$
    \E_{\bX \sim Q_\theta} \left[ \nabla_\theta \ln 
 Q_\theta(\bX) \right] 
    =\sumX Q_\theta(\bX) \nabla_\theta \ln Q_\theta(\bX)
    =\sumX \nabla_\theta Q_\theta(\bX)
    = \nabla_\theta \[\sum_{\bX} Q_\theta(\bX)\] = \nabla_\theta 1 = 0 \,,
$ is used to obtain the second line.} $R(\bX) \coloneqq \beta E(\bX) + \ln Q_\theta (\bX)$. 
When $R(\bX)$ is large, which corresponds to a higher system energy, the optimizer will tend to despise such configuration and hence reduce the variational free energy iteratively.
We emphasize that the score function depends on both the model distribution $Q$, and the target distribution $P$ through $E(\bX)$ resulting estimation variance.

\paragraph{Control Variance.}
That is, such a gradient estimator is noisy in practice and, the high variance gradients would require very small update steps, resulting in slow convergence.
To mitigate the variance, we subtract an $\bX$-independent but model-dependent\footnote{That is, the baseline $b$ depends on the underlying Ising model via the energy function $E(\bX)$.} constant baseline \cite{mnih2016variational,ranganath2013black}
\bea
    b \coloneqq \E_{\bX \sim Q_\theta} R(\bX) 
    = \E_{\bX \sim Q_\theta} \left[ \beta E(\bX) + \ln Q_\theta (\bX) \right] ,
    \label{eqn:baseline}
\eea
to $R(\bX)$ without affecting the estimation\footnote{This subtraction leaves the gradient estimator unbiased because $b$ amounts a $\bX$-independent term which has the expectation of $0$ over $Q$.},
\bea
\label{eqn:est_grad_b}
    \nabla_\theta \mathcal{F}_\theta = \frac{1}{\beta} \E_{\bX \sim Q_\theta} \left[ \nabla_\theta \ln Q_\theta(\bX) \cdot ( R(\bX) - b ) \right] \,.
\eea
Consequently, the objective function \eqref{eqn:free_energy} can be modified with the model-dependent variance reduction baseline,
\bea
    \mathcal{L} = \E_{\bX \sim Q_\theta} \left[ \beta E(\bX) + \ln Q_\theta (\bX) -b \right],
\eea
where $b$ is estimated at each iteration. 

\begin{algorithm}[h]
\caption{\code{CoRMF}}
\SetKwInOut{Input}{Input}
\Input{Ising model of size $|\bX|=N$ with structural constants $(J,h)$, criticality-ordered spin sequence $\bX_{i<j}$ from greedy Algorithm~\ref{algo:order}, learning rate $\eta$, max training steps $T$ and sample batch size $K$.}
\kwInit{Initialize an RNN neural network $Q_{\theta_0}$}
\For{$t=1,\cdots,T$}
{   
    \For{$k=1,\cdots,K$}
    {
        \For{$i=1,\cdots,N$}
        {
            Compute $Q_\theta(\bx^k_i|\bX^k_{j<i})$ to sample $\hat{\bx}^k_i$;\\
        }
        Put obtained sample  $\hat{\bX}^k=(\hat{\bx}^k_1, \cdots, \hat{\bx}^k_N)$ 
        into a sample set $\mathcal{S}_t$;\\
       
     } 
     Compute the energy $E(\bX)$ of each sample in $\mathcal{S}_t$;\\
     Compute the constant baseline $b$ according to \eqref{eqn:baseline};\\
     Estimate the gradient $\nabla_\theta \mathcal{F}_\theta$ according to \eqref{eqn:est_grad_b};\\
     Update RNN's parameter $\theta \gets \theta - \eta\nabla_\theta \mathcal{F}_\theta$
}
    \label{algo:algo_CoRMF}
\end{algorithm}

\section{Theoretical Analysis}

\label{sec:theory}
In this section, 
we first collect some results from Naive Mean-Field (NMF) in Sec.~\ref{sec:nmf}
, then introduce some useful analysis tools from matrix cut decomposition in Sec.~\ref{sec:matrix_cut}, and
lastly our main theoretical results in Sec.~\ref{sec:main_thm} which analytically characterize the variational energy $\calF_{\theta}^\star- \calF$ in \eqref{eqn:free_energy}.

To start with, for the generality of the main theorem, we provide the following lemma to convert any Ising model into its non-external-field formulation. 
\begin{claim}
\begin{lemma}[External Field Absorption \cite{griffiths1967correlations}]
\label{lemma:absorption}\vspace{-2mm}
For any Ising model $\calG=(V,E)$ of $N$ binary spins ($\bx_i=\{\pm1\}$) with non-zero external field $h$, $\calG$ can be represented as another \textit{extended} Ising model $\Tilde{\calG}$ of $(N+1)$ spins without external field by adding one auxiliary binary spin variable $\bx_{N+1}=\{\pm1\}$ connected to every node of the graph $\calG$. 
If strength matrices and external fields of the two models are then linked via the following relation
\bea
\tilde{J}_{ij}=
    \begin{cases}
        h_i, & \text{for } j=N+1\\
        0,  & \text{for } i=N+1\\
        J_{ij}, & \text{else }\label{lemma_equi}
    \end{cases}
\eea
where $\tilde{J}$ is a $(N+1)\times (N+1)$ \textit{asymmetric} matrix with $h$ as its last column and zeros as its last row, the free energies of the two Ising models are equivalent up to an additive constant.
\end{lemma}
\end{claim}
\begin{proof}
A detailed proof is shown in the Appendix~\ref{sec:pf_absorption}
\end{proof}
\begin{remark}
With Lemma~\ref{lemma:absorption}, we now ready to apply results of matrix cut decomposition \cite{jain2018mean,frieze1999quick} to prove our error bound.
For the ease of notation, from now on we denote $N\gets N+1$ and $1\gets\beta$ without loss of generality for the rest of theoretical analysis by keeping in mind that we are working with Ising models without external fields\footnote{Namely, we can cast any Ising model with external field into another Ising model with the same $\calF$ but with no external field.}.
\end{remark}

\subsection{Naive Mean-Field (NMF)}
\label{sec:nmf}

Here we recall the classical mean-field method, the Naive Mean-Field (NMF), and some of its useful results \cite{wainwright2008graphical}.
In NMF, the variational model family $Q_\theta$ is completely factorized, i.e.
$
Q(\bX) = \prod_i^N q_{\theta,i}(\bx_i),
$
and all spin variables are assumed independent.
From a graphical perspective, the NMF ansatz eliminates all edges between nodes, and the interactions are approximated and summarized in terms of background fields (i.e. \textit{mean}-field.)
Given NMF's simplicity, the minimization of \eqref{eqn:free_energy} is fairly easy to optimize, as can the inference.

By denoting $q_{i,\theta}(+1)\coloneqq p_{i,\theta}$ (and 
$q_{i,\theta}(-1)\coloneqq1-p_{i,\theta}$) the probability of the $i$'th spin being +1 (-1), the minimized NMF variational free energy takes the form
\begin{align}
\label{eqn:F_naive_MF}
\calF^\star_{\text{naive}}
\coloneqq
\calF_{\theta^\star,\text{naive}}
=&\underset{ \bar{X}\in [-1,1]^N}{\Min}
\Bigg[ 
\(\sum_{ij} J_{ij}\bar{x}_i\bar{x}_j+\sum_i h_i \bar{x}_i\)
\nonumber\\
&-\frac{1}{\beta}\sum_i H\(\frac{\bar{x}_i+1}{2}\) \Bigg],
\end{align}
where the mean-field value of the $i$th spin $\bar{x}_i$ is associated with $p_{i,\theta}$ via
$(2p_{i,\theta}-1)\equiv \bar{x}_i \in [-1,1]$, $H(\cdot)$ is the entropy, and the optimized mean-field values of $\calF^\star_{\text{naive}}$, $\bar{X}^\star\coloneqq \{\bar{x}_1^\star,\cdots,\bar{x}_N^\star\}$, are associated with the optimized variational parameters $\theta^\star$.
The derivation of \eqref{eqn:F_naive_MF} can be found in Appendix~\ref{sec:derivation_NMF}.

As an exemplar of VMF, it is easy to see that NMF solution \eqref{eqn:F_naive_MF} gives a valid upper bound for not only the variational free energy $\calF_\theta$, but also a upper bound for the true, yet intractable, free energy $\calF$.
To characterize the quality of this approximation, we firstly provide the following
lemma as an useful example, and later, as the main focus of this section, we aim to obtain tighter error bound for both NMF and CoRMF.
\begin{claim}\vspace{-2mm}
\begin{lemma}[Naive Bound for Naive Mean-Field]
\label{lemma:trivial_bound_naiveMF}
Following \eqref{eqn:naive_MF}, we have the following naive bound
\bea
\calF^\star_{\text{naive}}-\mathcal{F} \le N \norm{J}_F,
\eea
for naive mean-field with $\beta=1$, where $\norm{J}_F$ denotes the Forbenius norm.
\end{lemma}
\end{claim}
\begin{proof}
A detailed proof is shown in the Appendix~\ref{sec:naive_bound}.
\end{proof}
Lemma~\ref{lemma:trivial_bound_naiveMF} provides an error bound related to the size of the system and the magnitude of the couplings, and will soon be handy.
We call it a naive bound as it is obtained by considering the naive case where all $\bar{\bx}_i$ in \eqref{eqn:F_naive_MF} are zeros. For a more sophisticated and tighter NMF bound,  see \cite{jain2018mean}.

Although NMF has been shown highly accurate on certain class of Ising graphs \cite{huang2008statistical,ellis1978statistics}, its performs poorly on general Ising models as it fails to capture interactions between spins.
Therefore, more sophisticated VMF methods beyond NMF have been and can be proposed, and as well as tighter bounds \cite{wiegerinck2013variational,xing2012generalized,yedidia2003understanding,jordan1999introduction}.

\subsection{Matrix Cut Decomposition Machinaries}
\label{sec:matrix_cut}

Our goal here is to quantify the term $\calF_{\theta}^\star- \calF$ in \eqref{eqn:free_energy}, which measures the quality of the mean-field \eqref{AR_ansatz} in use.
Motivated by \cite{jain2018mean,borgs2012convergent}, we resort to another approximation scheme at graph level to quantify $\calF_{\theta}^\star- \calF$, the matrix cut decomposition, 
which intuitively allows us to approximate any Ising model as a collection of \textit{smaller} matrices of different sizes.
Each of them is associated with some cut-quantities that depend only on the edges in it.
With matrix cut, we shall show that the \textit{matrix-cut-approximated variational free energy}\footnote{Here, by approximated, we mean the cut-decomposed free energy. Note that ``variational free energy'' itself is already an approximation by mean-field method.}
itself is a variational ansatz, which can be interpreted as the dominant term of the partition function of another analytically tractable \textit{auxiliary Ising model}, see Fig.~\ref{fig:1} and \eqref{eqn:Z_D_variational} in Appendix~\ref{sec:pf_main_thm}.
We apply this  auxiliary Ising ansatz to construct $\calF_{\theta}^\star- \calF$.

We start with notations.
Following the \cite{frieze1999quick,jain2018mean, jain2017approximating}, we index the rows and columns of any $m\times n$ matrix $M$ by sets $[m]$ and $[n]$, where $[k]\coloneqq\{1,\cdots,k\}$ denotes a set of numbers up to $\abs{[k]}=k$.

\begin{claim}\vspace{-2mm}
\begin{definition}[Cut Matrix]
Given the column subset $[s]\subseteq[m]$, the row subset $[t]\subseteq [n]$ and a real value $d$, we define the $[m]\times [n]$ \textit{Cut Matrix} $\mathsf{CUT}([s],[t],d)$ by
\bea
\mathsf{CUT}_{ij}([s],[t],d) = 
\begin{cases}
    d\quad \text{if } (i,j)\in [s]\times [t],\\
    0\quad \text{otherwise}.
\end{cases}
\eea
\end{definition}
\begin{definition}[Cut Decomposition]
The width-$p$ \textit{Cut Decomposition} of a give matrix $A$ is defined by
\bea
A = D^{(1)} + D^{(2)} + \cdots +D^{(p)} + W,
\eea
where $D^{(\mu)} \coloneqq \mathsf{CUT}\([s]_\mu,[t]_\mu,d_\mu\) $ for $\mu=1,\cdots, p$.
We say such a decomposition has \textit{width} $p$, \textit{coefficient length} $\(\sum_{\mu=1}^p d_\mu^2\)^{\nicefrac{1}{2}}$ and \textit{error} $\norm{W}_C$.
\end{definition}
\end{claim}
Next, we introduce the following matrix norms: 
\footnotesize
\bea
\norm{M}_F &\coloneqq& \sqrt{\sum_{(i,j)\in [m]\times [n]} M_{ij}^2},\quad (\text{Forbenius Norm})\nonumber\\
\norm{M}_{\infty} &\coloneqq&
\underset{(i,j)\in [m]\times [n]}{\Max} \abs{M_{ij}},\quad  (\text{Maximum Norm})\nonumber\\
\norm{M}_C &\coloneqq& \underset{[s]\subseteq[m],[t]\subseteq [n]}{\Max} \abs{M\([s],[t]\)},\quad  (\text{Cut Norm})\nonumber\\
\|W\|_{\infty \mapsto 1} &\coloneqq& \underset{\|\bX\|_{\infty}\leq 1}{\Max}\|W\bX\|_{1},\quad (\text{Infinity-to-One Norm})\nonumber
\eea\normalsize
where $M\([s],[t]\)\coloneqq \sum_{(i,j)\in [s]\times [t]}M_{ij}$ and
$\|W\bX\|_{1}\coloneqq \sum_i\abs{\sum_j W_{ij}\bx_j}$.

With above, we introduce next lemma to bound the closeness of two distinct cuts on the same Ising model $J$.
\begin{claim}\vspace{-2mm}
\begin{lemma}[Cut Perturbation, Modified from  \cite{frieze1999quick,jain2018mean}]\label{lemma:E_decomposed_bound}
Following above definitions of $J, D^{1},\dots,D^{p}$, given real numbers $s_{\mu},s'_{\mu},t_{\mu},t'_{\mu}$ for each $\mu\in[p]$ and some 
$\gamma \in (0,1)$ such that $\abs{s_\mu},\abs{t_\mu},\abs{s'_\mu},\abs{t'_\mu} \le N$, 
$\abs{s_\mu - s'_\mu} \le \gamma N$ and $\abs{t_\mu - t'_\mu} \le \gamma N$ 
for all $\mu\in[p]$, we have
$-8\norm{J}_F\gamma N \sqrt{p}\le \sum_\mu d_\mu
\abs{r'_\mu c'_\mu - r_\mu c_\mu} \le 8\norm{J}_F\gamma N \sqrt{p}$. 
\end{lemma}
\end{claim}
\begin{proof}
A detailed proof is shown in the Appendix~\ref{sec:pf_E_decomposed_bound}.
\end{proof}

Lemma~\ref{lemma:E_decomposed_bound} enable us to bound the element-wise error resulting from cut-decomposing an Ising model $J$. 
Moreover, it characterizes how the changes in \textit{cut} affect the error bound, and hence allows us to construct the variational ansatz over different cuts.

Finally, we introduce next lemma to characterize the error in free energy introduced by cut decomposition.
\begin{claim}\vspace{-2mm}
\begin{lemma}[\cite{jain2018mean}]
\label{thm:naive_MF_bound} 
Let $J$ and $D$ be the matrices of interaction strengths and vectors of external fields of 
Ising models with partition functions $\calZ_{J}$ and $\calZ_{D}$, and variational free energies $\mathcal{F}_{J,\text{naive}}$ and $\mathcal{F}_{D,\text{naive}}$ of naive mean-field. 
Then, with $W:= J - D$, 
we have  
$
\left|\ln \calZ_J-\ln \calZ_{D}\right|\leq \abs{W}_{\infty \mapsto 1}$,
and 
$\left|\mathcal{F}^\star_{J,\text{naive}}-\mathcal{F}^\star_{D,\text{naive}}\right|\leq\|W\|_{\infty \mapsto 1}.$
\end{lemma}
\end{claim}
\begin{proof}
The detailed proof can be found in Appendix~\ref{appendix:them_NMF_bound} or original proof
\cite[Lemma~17]{jain2018mean}.
An extension with external field is also provided in Appendix~\ref{sec:naive_MF_bound_with_h}.
\end{proof}
\begin{remark}
For the ease of notation, we use the shorthand $\calF^\star_J=\mathcal{F}^\star_{J,\text{naive}}$ and $\calF^\star_D=\mathcal{F}^\star_{D,\text{naive}}$ in this paper.
\end{remark}

\subsection{Main Theorem}
\label{sec:main_thm}
We begin this section by pointing out a trivial lemma that connects the upper bounds of NMF and CoRMF variational free energies.
Then, we present our main theoretical result as Theorem~\ref{thm:main_thm}.

Given the fact that RNN is able to learn the naive mean-field factorization (completely factorized $Q_\theta$), we have:
\begin{claim}\vspace{-2mm}
\begin{lemma}
\label{lemma:F_RNN}
Any generalized mean-field $\calF_{\text{general}}$, that includes the naive mean-field as a member of its model family, has the following property
: $\abs{\mathcal{F}^\star_{\text{naive}}-\calF} \geq \abs{\calF^\star_{\text{general}}-\calF}.$

\end{lemma}
\end{claim}
\begin{remark}
Lemma~\ref{lemma:F_RNN} states that as the naive mean field belongs to the family of autoregressively factorized distributions, we have 
$\abs{\mathcal{F}^\star_{\text{naive}}-\calF} \geq \abs{\calF^\star_{\text{RNN}}-\calF}.$
\end{remark}

Now we can sate our main result: a provably error bound for CoRMF.
\begin{claim}\vspace{-2mm}
\begin{theorem}[Main Theorem]
\label{thm:main_thm}
Given an Ising model $J$ of $N$ nodes and its free energy $\calF$, the CoRMF approximates $\calF$ with $Q_\theta$ following the error bound
\bea\label{eqn:main_bound}
\mathcal{F}^\star_{\text{CoRMF}}- \mathcal{F}
\le 42 N^{\nicefrac{2}{3}} \norm{J}_F^{\nicefrac{2}{3}} \ln^{\nicefrac{1}{3}}\(48 N \norm{J}_F + e\),\nonumber
\eea
which is equivalent to a lower bound of the KL-divergence \eqref{eqn:KL}, $\textsf{KL}\(Q_{\theta}|P\)/\beta
$.
\end{theorem}
\end{claim}

\begin{proof}[Proof Sketch]
Since neural networks give only inaccessible black-box functions, it is hard to compute the error bound $\calF_{\text{RNN}}^\star-\calF$ for CoRMF directly.
To get around this, we start with constructing the NMF error bound
$\calF_{\text{naive}}^\star-\calF$, and then treat it as an inclusive special case of CoRMF (Lemma~\ref{lemma:F_RNN}) to obtain  $\calF_{\text{RNN}}^\star-\calF \le
        \calF^\star_{\text{naive}}-\mathcal{F}$
as the error bound for CoRMF.

To construct the NMF error bound $\calF^\star_{J,\text{naive}}-\mathcal{F}$,
    we first divide it into three terms
\begin{align}
    \calF^\star_{J,\text{naive}}-\mathcal{F}
        \leq& 
        \blue{|\ln \calZ -\ln \calZ_{D}|}
        +
        \Blue{|\calF^\star_{J,\text{naive}}-\mathcal{F}^\star_{D,\text{naive}}|
        }\nonumber\\
        &+
        \green{|\ln \calZ_D - \mathcal{F}^\star_{D,\text{naive}}|},
        \label{ineq_3terms}
\end{align}
    by inserting a $D$-cut free energy $\ln \calZ_D$ and its corresponding minimized NMF variational free energy $\calF^\star_{D,\text{naive}}$.
    Then we bound them separately, and finalize the proof with Lemma~\ref{lemma:F_RNN}. 
    The detailed proof consists  4 conceptual steps:
    
\paragraph{Step 1:}
        To bound \blue{$|\ln \calZ -\ln \calZ_{D}|$}, $\Blue{ |\calF^\star_{J,\text{naive}}-\mathcal{F}^\star_{D,\text{naive}}|}$,
        we use cut decomposition to find suitable $D$ and apply Lemma~\ref{thm:naive_MF_bound}.
        
\paragraph{Step 2:}
        To bound \green{$|\ln \calZ_D - \mathcal{F}^\star_{D,\text{naive}}|$}, we introduce another $\gamma$-parametrized variational ansatz $\ln \calZ_{D,\gamma}^\star$ 
        such that
\begin{align}
    \green{
    \abs{\ln \calZ_D - \mathcal{F}^\star_{D,\text{naive}}}
    }
    \le &
    \abs{\ln \calZ_{D} - 
    \ln \calZ_{D,\gamma}^\star}\nonumber\\
    &+
    \abs{\ln \calZ_{D,\gamma}^\star-\mathcal{F}^\star_{D,\text{naive}}},\nonumber
\end{align}
        where $|\ln \calZ_{D}- 
        \ln \calZ_{D,\gamma}^\star|$ and $|\ln \calZ_{D,\gamma}^\star-\mathcal{F}^\star_{D,\text{naive}}|$ can be controlled  by using cut perturbation Lemma~\ref{lemma:E_decomposed_bound}.
        Rationale of introducing the tractable ansatz $\ln \calZ_{D,\gamma}^\star$ is to approximate the intractable $\ln \calZ_D$ with controllable perturbation characterizing by $\gamma$.
        Mathematically, 
        $\calZ_{D,\gamma}$ is an auxiliary Ising model defined by approximating each configuration of $\calZ_{D}$ with error radius $\gamma N$;
        and $\calZ_{D,\gamma}^\star$ corresponds to its dominant term while other terms are exponentially suppressed, namely a saddle point approximation.
        Physically, $\calZ_{D,\gamma}^\star$ can be understood as a tractable \textit{degenerated single-state}\footnote{By degenerated single-state, we mean this Ising model has only one energy state, and can be multiple configurations in it.} Ising model that serves as a perturbed approximation capturing the dominant part of $\calZ_{D}$ such that Lemma~\ref{lemma:E_decomposed_bound} is applicable to provide an analytic error bound.

\paragraph{Step 3:}
        Combining above and applying Lemma~\ref{lemma:trivial_bound_naiveMF}, we arrive the bound for $|\calF^\star_{J,\text{naive}}-\mathcal{F}|$ in the form of \eqref{ineq_3terms}.

\paragraph{Step 4:}
        From above, we complete the proof by obtaining $\calF_{J,\text{CoRMF}}^\star-\calF$ via Lemma~\ref{lemma:F_RNN}.

    A detailed proof is shown in the Appendix~\ref{sec:pf_main_thm}.
\end{proof}

\section{Experimental Studies}
\label{sec:exp}
To validate numerically, we implement CoRMF with simplest RNN to perform forward Ising inference with given structural couplings $(J,h)$.
We demonstrate the efficiency and precision of our method on a variety of Ising models; and meanwhile showcase
how statistical quantities of interest can be estimated with CoRMF.
In particular, we evaluate the capability of CoRMF by comparing three distinct baselines\footnote{We emphasize again that, since there is not data/evidence in the problem setting, traditional variational inference methods are not applicable here.}:
(1) CoRMF with Random Order
(RO-), (2) CoRMF with Inverse Order (IO-), and (3) Naive Mean-Filed (NMF).

\subsection{Datasets}

\paragraph{N=100 1D Spin Chain.}
We first demonstrate the efficacy of CoRMF on the 1D Ising spin chain of size $N=100$,
$
E(\bX)=\sum_{\Braket{ij}} J_{ij} \bx_i\bx_j + \sumN h_i \bx_i,
$
where the spins are aligned in a 1D line and $\Braket{ij}$ denotes the spin-spin interaction $J_{ij}\bx_i\bx_j$ only exists between neighbor spins, i.e. $J_{ij}\bx_i\bx_j=0$ for all $j\neq i\pm 1$ for a given $i$.
In this nearest-neighbor interacting setting, we set $J_{ij}<0$ and $h_i>0$ such that all spins tend toward alignment with $-1$.

\paragraph{Dense N=10 Ising Model.}
Next, we use an Ising model of size N=10 with a slightly restricted interaction.
Specifically, the symmetric interaction strength matrix $J$ has $1,2,...,55$ as elements (with random assignment \textit{without replacement}) for the upper triangle except the diagonals, and we set $h_i>0$ for all $i$. 
In such Ising model, no ambiguity exists in the generated criticality spin order from Sec.~\ref{sec:critical_order} since there is no repeated $|J_{ij}|$ except $|J_{ji}|$ (recall $J$ is symmetric.)
Thus, this setup, differed from commonly used uniform $J$ (i.e. $J_{ij}=J$ for all $(i,j)$), nearest-neighbor \& random $J$ (e.g. Edwards–Anderson model \cite{edwards1975theory} ), or  general random $J$ (e.g. Sherrington-Kirkpatrick  \cite{sherrington1975solvable} or Hopfield \cite{hopfield1982neural} spin glasses), is mainly to verify the effectiveness of the criticality spin order.
We also examine two distinct temperatures (or equivalently rescaling $J$) for this model to see how CoRMF adapts to randomness\footnote{The distribution would become more random as the temperature rose.}.

\DeclarePairedDelimiter\floor{\lfloor}{\rfloor}
\paragraph{Dense \& Sparse \& Random N=20 Ising Models.}
Then, we use a dense Ising model with $N=20$, where $J_{ij} \sim \calU([L])$ (uniformly pick from $[L]$ with some number $L$) is dense and the ambiguity of criticality-ordered sequence can be controlled by tuning $L$.
We test CoRMF on  $L=N^2$ (rarely ambiguous) and $L=\floor{\sqrt{N}}$ (highly ambiguous) to see how CoRMF is affected by order ambiguity.
Further, we consider a sparse Ising model with $N=20$, where $J_{ij} \sim \mathsf{Poisson}(\lambda=0.4)$ is sparse and the criticality-ordered sequence is ambiguous due to repeated values (and sparsity.)
Last, a setting with general interaction $J_{ij} \sim \calU([L=5]-2)$ is conducted.
For all N=20 Ising models, we set $h=0$.
These settings aim to mimic real-world forward Ising problems where $J$ is complex, possibly sparse and the order is easily contaminated.

\subsection{Tasks: Forward Ising Inference}

We consider two forward Ising inference tasks: computing the minimized variational free energy, and the magnetization (global mean). 

\paragraph{Minimized Variational Free Energy $\calF^\star$.}
For CoRMFs, we report the average values of the variational free energy of RNN samples as the minimized variational free energy $\calF^\star$.
For NMF, the $\calF^\star$'s are minimized according to \eqref{eqn:naive_MF}.

\paragraph{Magnetization $\Braket{\bx}_Q=\sum_i\E_Q(\bx_i)$.} 
The magnetization of an Ising model is the mean of mean parameters of spins
\footnotesize
\bea
\Braket{\bx}_Q
\coloneqq\frac{1}{N}\sum_i\E_{x\sim Q}(\bx_i)
=\frac{1}{N}\sum_i\[2q_{\theta^\star}(\bx_i=+1)-1\].\nonumber
\eea\normalsize
We compute and report magnetization values for both CoRMF and baselines.
To evaluate estimated $\Braket{\bx}_Q$ easily, we introduce external field $h$ in part of datasets to force $\Braket{\bx}_Q$ deviate from zero.

For all these tasks, we use samples obtained with Gibbs MCMC sampler to compute reference values. 
For each datasets, we repeat all above tasks 5 times for CoRMFs and 10 times for NMF, report the results in Table~\ref{tab:results}, and summarize implementation details in Appendix~\ref{sec:exp_details}.

\begin{table}[h]
\caption{Comparison of CoRMF and NMF. 
We examine the effectiveness of the proposed relative order by considering criticality-ordered, random-ordered (RO-), inverse-ordered (IO-) CoRMFs, comparing with NMF.}
\label{tab:results}
\resizebox{0.48\textwidth}{!}{
\begin{tabular}{llllll}
\toprule
Dataset & 
Mean Field &
 $\calF^\star $ &
 $\Braket{\bx}_Q$ 
 \\ 
\midrule
 & CoRMF & \textbf{-300.00489}$\pm$0.00022 & \textbf{-0.99985}$\pm$0.00004\\
  & RO CoRMF & \textbf{-300.00511}$\pm$0.00018 & \textbf{-0.99985}$\pm$0.00002\\
{N=100 1D Spin Chain} & IO   CoRMF & \textbf{-300.00509}$\pm$0.00028& \textbf{-0.99986}$\pm$0.00002\\
    & NMF  & -292.50790$\pm$0.72770 & -0.96880$\pm$0.00250 \\
    & Reference  & - &  -0.99989$\pm$0.00156   \\
\midrule
 & CoRMF & \textbf{-85.34812}$\pm$0.00013  &\textbf{-0.09493}$\pm$0.00135 \\
  & RO CoRMF & -85.34548$\pm$0.00038 & -0.09592$\pm$0.00118 \\
{N=10 Ising ($\beta$=1)}  & IO   CoRMF & -85.34647$\pm$0.00271  & -0.09530$\pm$0.00118\\
    & NMF  & -74.83870$\pm$6.88050 & -0.11750$\pm$0.06920 \\
   & Reference  & - & -0.08862$\pm$0.09935 \\
\midrule
& CoRMF  & \textbf{-423.91318}$\pm$0.00002&\textbf{-0.08027}$\pm$0.00047\\
  & RO CoRMF  & -423.00000$\pm$0.00000& -0.20000$\pm$0.00000 \\
N=10 Ising ($\beta$=5)& IO CoRMF & -423.39996$\pm$0.00000& 0.00000$\pm$0.00000 \\
    & NMF  & -369.35510$\pm$37.30090 & -0.12390$\pm$0.07100 \\
   & Reference  & - & -0.06342$\pm$0.09307   \\
   \midrule
 & CoRMF & \textbf{-166.06870}$\pm$0.00550  & \textbf{-0.00016}$\pm$0.00157\\
  & RO CoRMF & -165.17768$\pm$6.55424 & 0.01161$\pm$0.02440 \\
{Dense N=20 Ising ($L$=400)}  & IO CoRMF & -164.15070$\pm$5.65829 & -0.00002$\pm$0.00107 \\
    & NMF  & -140.20250$\pm$14.3210 & -0.00450$\pm$0.03950 \\
   & Reference  & -&     -0.00025$\pm$0.05324  \\
   \midrule
 & CoRMF & \textbf{-94.72956}$\pm$0.00123 & \textbf{0.00008}$\pm$0.00023\\
  & RO CoRMF & -94.72362$\pm$0.00998 & 0.00006$\pm$0.00020 \\
{Dense N=20 Ising ($L$=5)}  & IO CoRMF & -93.43296$\pm$2.52553 & -0.00037$\pm$0.00075\\
    & NMF  & -81.11440$\pm$5.19200 & -0.00430$\pm$0.02760  \\
   & Reference  & -&   0.00008$\pm$0.014966 \\
  \midrule
 & CoRMF & \textbf{-78.81788}$\pm$0.00169 & \textbf{0.00021}$\pm$0.00109 \\
  & RO CoRMF &  -67.45300$\pm$0.59780 &  -0.00737$\pm$0.00907 \\
Sparse N=20 Ising   & IO CoRMF &-78.81728$\pm$0.00022 & 0.00064$\pm$0.00181\\
    & NMF  & -66.28090$\pm$5.79250 &-0.03010$\pm$0.06360  \\
  & Reference  & - & -0.00165$\pm$0.09725\\
  \midrule
 & CoRMF & -149.38675$\pm$0.00013 & 0.00081$\pm$0.00036 \\
  & RO CoRMF &  \textbf{-149.38687}$\pm$0.00013 & 0.00053$\pm$0.00134 \\
Random N=20 Ising   & IO CoRMF &-141.12375$\pm$11.31409 & \textbf{-0.00039}$\pm$0.00102\\
    & NMF  & -130.62220$\pm$9.50900 & -0.00780$\pm$0.07020  \\
  & Reference  & - & -0.00143$\pm$0.0729\\
\bottomrule
\end{tabular}}
\end{table}

\subsection{ Forward Ising Inference Quality}
In our implementation, CoRMF agrees with Theorem~\ref{thm:main_thm} by outperforming all baselines in minimizing $\calF^\star$ on all datasets except N=20 with random $J$; when estimating magnetization in cases with non-zero external field $h$, our method was at least $\sim$4+\% more accurate than NMF baselines; however CoRMF failed to excel on highly sparse and ambiguous graphs.

In \textit{N=100 1D Spin Chain} experiments, the CoRMF outperformed NMF in both $\calF^\star$ and $\Braket{\bx}_Q$ regardless which spin order was used, which is not surprising as the model is highly restricted and simple.
In \textit{N=10 Ising Model} experiments, the results show that not only did CoRMF outperform NMF, but criticality-ordered was superior to RO- \& IO-;
moreover, in the low-temperature setting ($\beta=5$), the significance of the tree order appears to increase. 
In \textit{Dense N=20 Ising Model} experiments, the results show that CoRMF triumphed in general when order was rarely ambiguous; even in the case of highly repeated $J$, CoRMF succeeded.
However, in \textit{Sparse \& Random N=20 Ising Models} experiments, the distinctions between CoRMF, RO-, and IO-CoRMF began to blur as the order became contaminated, although CoRMF was still superior by a small margin on average.

\section{Conclusion}
\label{sec:conclusion}
In this paper, we propose a neutralized tree-structured variational mean-field, CoRMF, as an efficient solver for forward Ising inference problems. 
Our framework is generic as (i) it is independent of the Ising model of interest, (ii) it makes no strong assumptions about the network architecture, and (iii) it is computationally efficient, expressive enough, and effortless to employ to any forward Ising inference problem.
Analytically,
we provide an approximation error bound for CoRMF.
Numerically, we conduct a variety of experiments that verify our theoretical results and CoRMF delivers on average 4+\% improvement in accuracy over baselines.
This work serves as an attempt towards developing deep-learning enhanced solvers for forward Ising problems. 
Nevertheless, there is one limitation we wish to highlight: instead of finding an optimal order, it adopts a heuristic-based order without theoretical guarantees, and therefore may fail to excel on certain graphs, see Appendix~\ref{sec:limitations} for more discussions.
For future investigations, our plan is two-fold: 
(1) relaxing CoRMF with the theoretically guaranteed information-distance-weighted tree \cite{choi2011learning,lake1994reconstructing} for both forward and inverse Ising problems in data-driven settings; 
(2) integrating the tree-structured mode finding algorithms like \cite{chen2014mode} with CoRMF.

\vspace{-2mm}
\section*{Acknowledgments}
We thank the authors of \cite{jain2018mean, nguyen2017inverse} for their prompt clarifications regarding their papers. Special thanks to Yu-Hsien Kung for enlightening discussions on related topics in the early stages of this work. We also extend our gratitude to the authors of \cite{reneau2023feature} for sharing their dataset and to anonymous referees for their constructive comments. EJK is grateful to the National Center for Theoretical Sciences of Taiwan for funding (112-2124-M-002-003).

\clearpage
\normalsize
\setlength{\parskip}{0.5em}
\appendix
\label{sec:append}
\onecolumn
\aistatstitle{Supplementary Materials}

\section{Technical Details}

\subsection{Algorithm for Generating the Criticality-ordered Spin Sequence}\label{sec:greedy_algo}

To order the importance of spins $(\bx_{1},\bx_{2},\cdots,\bx_{n})$, we construct the maximum spanning tree of the Ising graph via a modified (and negated) Kruskal's algorithm \cite{graham1985history}. 
A maximum spanning tree is a spanning tree of a given  weighted graph $\calG$, whose sum of weights is maximal. 
One method for computing the maximum weight spanning tree can be summarized as follows.

\begin{enumerate}
    \item Sort edges of $\calG$ into non-increasing order by weight $\abs{J_{u,v}}$.
    Note that, there might be equally weighted edges and hence the resulting tree is not always unique.
    In case of equally weighted edges, we  always sort them arbitrarily.
    Therefore, the obtained spin order might be contaminated with ambiguities.
    
    \item 
    Let $B$ be the set of edges consisting the tree with the maximum weight spanning, and $A$ be the relative order (a list) among spin associated to $B$. 

    \item
    Initialize $A= \emptyset$ and $B = \emptyset$. 
    
    \item 
    Add the first edge $(u,v)$ to $B$ and append $u$ and $v$ into list $A$.
    
    \item Add the next edge $(u,v)$ to $B$ if and only if it does not form a cycle in $B$. 
    There are two possible cases:
    (i) one of the node (either $u$ or $v$) is already in $A$, then we just append another to $A$;
    (ii) both nodes $u,v$ are not in our $A$, then we append them randomly to $A$.
    
    \item Keep adding edges (step 4) until $B$ has $\abs{\calV}-1$ edges then return $A.$
\end{enumerate}

\begin{algorithm}[h]
\caption{\code{Criticality-Weighted Maximum Spanning Tree}}
\label{algo:order}
\SetKwInOut{Input}{Input}
\Input{$\calG = (\calE,\calV)$ a connected  undirected graph, weighted by $J$}
\kwInit{a union-find data structure $B\gets \emptyset $, an empty list
$A\gets \emptyset $} 
\For{each vertex $v \in \calV$
    }{
    $\mathsf{MAKE\_SET}(v)$ 
    \blue{\footnotesize\Comment*{$\mathsf{MAKE\_SET}(v)$ puts $v$ in a set by itself}}
    }
sorted edge into  non-increasing order by absolute value of weight $\abs{J_{u,v}}$\\
\For{each edge $e_{u,v}=(u,v) \in \calE$ taken from above sorted list}{
    \If{$\mathsf{FIND\_SET}(u)$ $\neq$ $\mathsf{FIND\_SET}(v)$
    }{ 
    $B \gets B \cup {(u, v)} \cup {(v, u)}$\blue{\footnotesize\Comment*{$\mathsf{FIND\_SET}(v)$ returns the name of $v$'s set}}
    $\mathsf{UNION}(\mathsf{FIND\_SET}(u)$, $\mathsf{FIND\_SET}(v)$)\blue{\footnotesize\Comment*{$\mathsf{UNION}(u,v)$ combines sets that $u$ and $v$ are in}}
    \uIf{$(u,v) \not\in A$} 
    {
    append $u$ and $v$ to $A$
    }
    \uElseIf{$u \in A, v \not\in A$}{append $v$ to $A$}
    \Else{append $u$ to $A$
    \blue{\footnotesize\Comment*{if $v \in A, u \not\in A$}}
    }
    }
    }
    \Return the criticality-ordered spin sequence $A=[\bx_{1},\bx_{2},\cdots,\bx_{n}]$ 
\end{algorithm}
For a graph with $\abs{\calE}$ edges and $\abs{\calV}$ vertices, Kruskal's algorithm can be shown to run in $\calO(\abs{\calE} \ln \abs{\calE})$ time, or equivalently, $\calO(\abs{\calE} \ln \abs{\calV})$ time, all with simple data structures.    

\section{Limitations}
\label{sec:limitations}
We discuss CoRMF's limitations here.
    As mentioned previously, the proposed order is sorted based on a heuristic definition for mission criticality (weight):
    $\abs{J_{u,v}}$, which in general lack of theoretical guarantees and quantification for error caused by the tree approximation.
    In fact, find an optimal tree/order itself is an NP-hard problem in our not-data-driven setting --- there are $d!$ possible permutations for the spin order.
    To the best of our knowledge, unlike the data-driven setting where we can use information distance \cite{lake1994reconstructing} as graph weight and obtain the optimal tree with  spanning tree algorithm 
    \cite{choi2011learning}, there is no known data-free solution to this problem.
    Therefore, we employ a heuristic in this work which has been empirically validated.
    More sophisticated (but possible ad-hoc) extensions against sparse and ambiguous graphs can be easily employed, such as a high-assumption weighting mechanism for the greedy algorithm based on the sample-free analytic expression for pairwise correlation \cite{nikolakakis2021predictive} or test-based heuristic weightings (i.e. sparsity test, ambiguity test, $J$-$h$ ratio test...e.t.c.).

\section{Supplemental Theoretical Results}

\subsection{Naive Mean-Field Variational Free Energy}
\label{sec:derivation_NMF}
The following is an useful expression for minimizing the variational free energy of the naive mean-field $Q_\theta(\bX)=\prod_{i=1}^{N} q_{i,\theta}(\bx_i)$.

Starting from the variational mean-field free energy
\bea
\calF_{\theta}=
\E_{Q_\theta}\[E(\bX) + \frac{1}{\beta}\ln Q_{\theta}(\bX)\],\label{free_energy}
\eea 
by denoting $q_{i,\theta}(+1)\coloneqq p_{i,\theta}$ and 
$q_{i,\theta}(-1)\coloneqq1-p_{i,\theta}$, the first term on the RHS gives
\bea
\E_{Q_\theta}\[E(\bX)\]&=& \E_{Q_\theta}\(\sum_{ij}J_{ij}\bx_i\bx_j+\sum_i h_i\bx_i\)
=\sum_{ij}J_{ij}\;\E_{q_{i,\theta}} (\bx_i)\E_{q_{j,\theta}} (\bx_j)+\sum_i h_i \;\E_{q_{i,\theta}} (\bx_i) \nonumber \\
& =& \sum_{ij}J_{ij}\(2p_{i,\theta}-1\)\(2p_{j,\theta}-1\)+\sum_i h_i (2p_{i,\theta}-1).
\eea 
As for the second term, recalling the entropy $H(\cdot)\coloneqq - \sum P(\cdot)\ln P(\cdot)$ , we have
\bea
\E_{Q_\theta}\[\ln Q_{\theta}(\bX)\]&=&\sum_i \E_{q_{i,\theta}} \[\ln q(\bx_i)\]\nonumber\\
&=&\sum_i p_{i,\theta} \ln p_{i,\theta}+(1-p_{i,\theta})
\ln \(1-p_{i,\theta}\)
=-\sum_i H\(p_{i,\theta}\).
\eea
Thus, we can rewrite the variational free energy of naive mean-field as
\bea
\calF_{\theta,\text{naive}}(\bar{x}_1,\bar{x}_2,\cdots,\bar{x}_n)=
 \sum_{ij}J_{ij}\bar{x}_i\bar{x}_j+\sum_i h_i \bar{x}_i-\frac{1}{\beta}\sum_i H\(\frac{\bar{x}_i+1}{2}\),
\eea 
by associating the mean-field value of the $i$th spin with $p_{i,\theta}$ via
$(2p_{i,\theta}-1)\equiv \bar{x}_i \in [-1,1].$ 
Therefore, we arrive a particular form of the minimized NMF variational free energy 
\bea 
\label{eqn:naive_MF}
\boxed{
\calF^\star_{\text{naive}}
\coloneqq
\calF_{\theta^\star,\text{naive}}
=\underset{ \bar{X}\in [-1,1]^N}{\Min}
\left[ 
\(\sum_{ij} J_{ij}\bar{x}_i\bar{x}_j+\sum_i h_i \bar{x}_i\)
-\frac{1}{\beta}\sum_i H\(\frac{\bar{x}_i+1}{2}\) \right ],
}
\eea 
where the optimized mean-field values of $\calF^\star_{\text{naive}}$, $\bar{X}^\star\coloneqq \{\bar{x}_1^\star,\cdots,\bar{x}_N^\star\}$, are associated with the optimized variational parameters $\theta^\star$.

\label{sec:appendix_proof}

\subsection{Lemma~\ref{lemma:E_decomposed_bound}'s Corollaries}
\begin{corollary}
\label{coro:dst_parity}
Lemma~\ref{lemma:E_decomposed_bound} is invariant under the transformation $d_\mu\leftrightarrow -d_\mu$, namely
$-8\norm{J}_F\gamma N \sqrt{p}\le \sum_\mu -d_\mu
\abs{r'_\mu c'_\mu - r_\mu c_\mu} \le 8\norm{J}_F\gamma N \sqrt{p}$ is also true.
\end{corollary}
\begin{proof}[Proof of Corollary~\ref{coro:dst_parity}.]
Omitted.
\end{proof}
\begin{corollary}
\label{coro:exp_dst}
Since $
\exp{\sum_{\mu=1}^p -d_\mu  {s_\mu t_\mu}}
\leq 
\exp{\sum_{\mu=1}^p -d_\mu  {s'_\mu t'_\mu}}
\exp{8\sqrt{p}\norm{J}_F\gamma N}
$,
we can simply have
$
\exp{\sum_{\mu=1}^p -d_\mu  {s_\mu t_\mu}}
\leq
 \exp{8\sqrt{p}\norm{J}_F\gamma N}
\exp{-\sum_{\mu=1}^p d_\mu  {s'_\mu t'_\mu}}
$.
\end{corollary}
\begin{proof}[Proof of Corollary~\ref{coro:exp_dst}]
Starting with Lemma~\ref{lemma:E_decomposed_bound}, we have
\bea
\exp{\sum_{\mu=1}^p -d_\mu \( {s_\mu t_\mu} - s'_\mu t'_\mu\) }
\leq
\exp{\sum_{\mu=1}^p -d_\mu \abs{ {s_\mu t_\mu} - s'_\mu t'_\mu} }
\leq 
\exp{8\sqrt{p}\norm{J}_F\gamma N},
\eea
which implies 
\bea
\exp{\sum_{\mu=1}^p -d_\mu  {s_\mu t_\mu}}
\leq 
\exp{\sum_{\mu=1}^p -d_\mu  {s'_\mu t'_\mu}}
\exp{8\sqrt{p}\norm{J}_F\gamma N}.
\eea
Therefore,
\bea
\exp{\sum_{\mu=1}^p -d_\mu  {s_\mu t_\mu}}
&=&\exp{\sum_{\mu=1}^p -d_\mu ({s_\mu t_\mu-s'_\mu t'_\mu}) }
\exp{-\sum_{\mu=1}^p d_\mu  {s'_\mu t'_\mu}}\nonumber \\
&\leq&\exp{\sum_{\mu=1}^p -d_\mu \abs{{s'_\mu t'_\mu-s_\mu t_\mu}} }
\exp{-\sum_{\mu=1}^p d_\mu  {s'_\mu t'_\mu}}\nonumber \\
&\leq&  \exp{8\sqrt{p}\norm{J}_F\gamma N}
\exp{-\sum_{\mu=1}^p d_\mu  {s'_\mu t'_\mu}}.
\eea
\end{proof}

\subsection{Lemma~\ref{lemma:trivial_bound_naiveMF}}
\label{sec:naive_bound}

\begin{proof}[Proof of Lemma~\ref{lemma:trivial_bound_naiveMF}]
According to \eqref{eqn:naive_MF}, we have
$\calF^\star_{\text{naive}} \leq -N \ln2$ by naively plugging all $\bar{x}_i=0, \forall i$. 
As for $\calF$, we have
\bea
\calF=-\ln (\sum_{\bX}\exp{-\sum_{ij}J_{ij}\bx_{i}\bx_{j}}) \geq -\ln \(2^N \exp{N \norm{J}_F}\)
=-N\ln2-N\norm{J}_F.
\eea
Therefore we arrive the naive bound
$\calF^\star_{\text{naive}}-\mathcal{F} \le N \norm{J}_F.$
\end{proof}

\subsection{Lemma~\ref{lemma:absorption}}
\label{sec:pf_absorption}
\begin{proof}[Proof of Lemma~\ref{lemma:absorption}]
We first write down the energy function of $\tilde{\calG}$ by adding an auxiliary spin $\bx_{N+1}=\{\pm1\}$ to $\calG$ following \eqref{lemma_equi}:
\begin{align}
\tilde{E}(\bX,\bx_{N+1})
=\sum_{i,j=1}^{N+1} \tilde{J}_{ij}\bx_i\bx_j 
=\sum_{i,j=1}^{N}J_{ij}\bx_i\bx_j+\(\sum_{i=1}^{N} h_i\bx_i\)\bx_{N+1}.
\end{align}
On the basis of which, the corresponding partition function can be expressed as
\begin{align}
\tilde{\calZ}&=\sum_{\bX; \bx_{N+1}\in \{\pm1\} } \exp{-\beta \tilde{E}(\bX)}  \\
& =\sum_{\bX} \exp{-\beta \(\sum_{i,j=1}^{N}J_{ij}\bx_i\bx_j+\sum_{i=1}^{N} h_i\bx_i\)}+\sum_{\bX} \exp{-\beta \(\sum_{i,j=1}^{N}J_{ij}\bx_i\bx_j-\sum_{i=1}^{N} h_i\bx_i\)} \nonumber \\
& = 2 \sumX \exp{-\beta \(\sum_{i,j=1}^{N}J_{ij}\bx_i\bx_j+\sum_{i=1}^{N} h_i\bx_i\)}
=2 \calZ, \annot{(Ising Parity Symmetry)}
\end{align}
indicating the free energies $\calF=-\ln \calZ / \beta$ of the two Ising models are equivalent up to an additive logarithmic-constant.
\end{proof}

\subsection{Lemma~\ref{lemma:E_decomposed_bound}}
\label{sec:pf_E_decomposed_bound}

To prove Lemma~~\ref{lemma:E_decomposed_bound}, we need the following Cut Decomposition Lemma. 
\begin{claim}
\begin{lemma}[Cut Decomposition \cite{frieze1999quick,jain2018mean}]
\label{lemma:Frieze_Kannan}
Given an arbitrary real $m\times n$ matrix $J$, and $\epsilon>0$.
We can find a cut decomposition of width at most $16/\epsilon^2$, coefficient length at most $4\norm{J}_F/\sqrt{mn}$, error at most $4\epsilon\sqrt{mn}\norm{J}_F$, and such that $\norm{W}_F\leq \norm{J}_F$.
\end{lemma}
\end{claim}
\begin{proof}[Proof of Lemma~\ref{lemma:Frieze_Kannan}.]
A detailed proof  can be found in \cite[Sec.~4.3]{frieze1999quick}.
Note that we rescale $\epsilon$ following \cite{jain2018mean} for simplicity (i.e. when proving inequalities, equal-$\epsilon$ weights are preferred\footnote{For example, in \eqref{ineq_3terms}, each $\abs{\cdot}$ on the RHS is proportional to $\epsilon/3$.}.)
\end{proof}
We are now ready to prove  Lemma~~\ref{lemma:E_decomposed_bound}.
\begin{proof}[Proof of Lemma~\ref{lemma:E_decomposed_bound}]

Using the argument of \cite[Lemma~19]{jain2018mean},
since $$ \abs{s'_\mu t'_\mu - s_\mu t_\mu} \le \abs{t'_\mu||s'_\mu - s_\mu} + \abs{t_\mu}\abs{s'_\mu - s_\mu} \le 2\gamma N^2,$$
we have
\begin{align*}
\(\sum_{\mu=1}^p d_\mu\abs{s'_\mu t'_\mu - s_\mu t_\mu}\)^2
&\le \(\sqrt{p}\cdot \sqrt{\sum_\mu^p d_\mu^2}\cdot  2\gamma N^2\)^2 \annot{(By Cauchy-Schwartz inequality)}
\\
&\le \(8\sqrt{p}\norm{J}_F\gamma N \)^2,
\annot{($\sqrt{\sum_\mu^p d_\mu^2}\le 4\norm{J}_F/N$ from Lemma~\ref{lemma:Frieze_Kannan})}
\end{align*} 
which implies
$-8\norm{J}_F\gamma N \sqrt{p}\le \sum_\mu d_\mu
\abs{r'_\mu c'_\mu - r_\mu c_\mu} \le 8\norm{J}_F\gamma N \sqrt{p}$.
\end{proof}

\subsection{Lemma~\ref{thm:naive_MF_bound}}
\label{appendix:them_NMF_bound}
\begin{proof}[Proof of Lemma~\ref{thm:naive_MF_bound}]
\cite[Lemma~17]{jain2018mean} provides the original proof and we restate it here only for self-containedness.
For any spin configuration $\bX\in\{\pm1\}^{N}$, we have 
\begin{align}
\abs{E_J(\bX) -E_D(\bX)}
&=\left|\(\sum_{i,j}J_{ij}\bx_i\bx_j\)-\(\sum_{i,j}D_{ij}\bx_i\bx_j\)\right| \\
&\leq
\left|\sum_{i}\(\sum_{j}W_{ij}\bx_j\)\bx_i\right| \annot{(\text{Cauchy-Schwartz inequality)}}\nonumber\\
& \leq 
\left|\sum_{i}\(\sum_{j}W_{ij}\bx_j\)\right| \leq\|W\|_{\infty\mapsto1},
\end{align}
which implies
\begin{equation}
\exp(-\sum_{i,j}J_{ij}\bx_i\bx_j) \in \left[ \exp\left(-\sum_{i,j}D_{ij}\bx_i\bx_j \pm \|W\|_{\infty \mapsto 1} \) \right],
\end{equation}
and 
$\left|\mathcal{F}^\star_{J,\text{naive}}-\mathcal{F}^\star_{D,\text{naive}}\right|
\leq\|W\|_{\infty \mapsto 1}$ 
according to \eqref{eqn:naive_MF}.
Taking the sum over all configurations
and take the logarithm on both sides, we arrive 
\bea
\ln\calZ_J=\ln\[\sumX \exp(-\sum_{i,j}J_{ij}\bx_i\bx_j) \]
\in
\ln \left[ \sumX \exp\left(-\sum_{i,j}D_{ij}\bx_i\bx_j \)  \right]\pm \|W\|_{\infty \mapsto 1},
\eea
which implies $\abs{\ln\calZ_J-\ln \calZ_D}\le \|W\|_{\infty \mapsto 1}$.
An extension of this theorem with external fields is provided in the following section.
\end{proof}

\subsection{An Extension of Lemma~\ref{thm:naive_MF_bound} with External Fields}
\label{sec:naive_MF_bound_with_h}

\begin{claim}
\begin{theorem}[Modified from \cite{jain2018mean}]
\label{thm:naive_MF_bound_with_h} 
Let $(J,h)$ and $(D,u)$ be the matrices of interaction strengths and vectors of external fields of 
Ising models with partition functions $\calZ_P$ and $\calZ_{Q}$, and variational free energies $\mathcal{F} $ and $\mathcal{F}^{*}_{Q}$. 
Then, with $W:= J - D$. 
We have  
\bea
\left|\log \calZ-\log \calZ_{Q}\right|\leq \|W\|_{\infty \to 1}+\sqrt{n}\|h-u\|_2,
\eea
and 
\bea
\left|\mathcal{F}^{*}-\mathcal{F}^{*}_{D}\right|\leq\|W\|_{\infty \to 1}+\sqrt{n}\|h-u\|_2.
\eea
\end{theorem}
\end{claim}
\begin{proof}[Proof of Theorem \ref{thm:naive_MF_bound_with_h}]
For any spin configuration $\bX\in\{\pm1\}^{N}$, we have 
\begin{align}
&\left|\(\sum_{i,j}J_{ij}\bx_i\bx_j+\sum_{i}h_i\bx_i\)-\(\sum_{i,j}D_{ij}\bx_i\bx_j+\sum_{i}u_i \bx_i\)\right| \\
&\leq
\left|\sum_{i}\(\sum_{j}W_{ij}\bx_j\)\bx_i\right|+\left|\sum_{i} \(h_i-u_i\)\bx_i\right| \annot{(\text{Cauchy-Schwartz inequality)}}\nonumber\\
& \leq 
\left|\sum_{i}\(\sum_{j}W_{ij}\bx_j\)\right|
+\sqrt{n}\norm{h-u}_2 \leq\|W\|_{\infty\mapsto1}+\sqrt{n}\|h-u\|_2.
\end{align}
From above we get $|\mathcal{F}^{*}-\mathcal{F}^{*}_{D}|\leq\|W\|_{\infty \to 1}$. 
Moreover, for any $\bX\in\{\pm1\}^{N}$, we have

\begin{equation}
\exp(\sum_{i,j}J_{ij}\bx_i\bx_j) \in \left[ \exp\left(\sum_{i,j}D_{ij}\bx_i\bx_j\) \pm \(\|W\|_{\infty \mapsto 1}+ \sqrt{n}\|h-u\|_2 \) \right].
\end{equation}
Taking first the sum of these inequalities over all $x\in\{\pm1\}^{n}$
and then the log, we get 
$|\mathcal{F}^{*}-\mathcal{F}^{*}_{D}|\leq\|W\|_{\infty \to 1}+||h-u||_2\sqrt{n}$. 
\end{proof}

\section{Proof of Theorem~\ref{thm:main_thm}}
\label{sec:pf_main_thm}

\begin{proof}[Proof of Theorem~\ref{thm:main_thm}]

We start with constructing the naive mean-field error bound
$\calF_{\text{naive}}^\star-\calF$, and then treat it as an inclusive special case of CoRMF (Lemma~\ref{lemma:F_RNN}) to obtain  $\calF_{\text{RNN}}^\star-\calF \le
        \calF^\star_{\text{naive}}-\mathcal{F}$
as the error bound for CoRMF.
To construct the naive mean-field error bound
    $\calF^\star_{J,\text{naive}}-\mathcal{F}$,
    we first divide it into three terms
\begin{align}
        \calF^\star_{J,\text{naive}}-\mathcal{F}
        \leq
        \blue{|\ln \calZ -\ln \calZ_{D}|}
        +
        \Blue{|\calF^\star_{J,\text{naive}}-\mathcal{F}^\star_{D,\text{naive}}|
        }
        +
        \green{|\ln \calZ_D - \mathcal{F}^\star_{D,\text{naive}}|},
\end{align}
    by inserting a $D$-cut free energy $\ln \calZ_D$ and its corresponding minimized NMF variational free energy $\calF^\star_{D,\text{naive}}$.
Then, our technical proof consists four conceptual steps:
\begin{itemize}\setlength\itemsep{0em}
        \item \textbf{Step 1:}
        To bound \blue{$|\ln \calZ -\ln \calZ_{D}|$}, $\Blue{ |\calF^\star_{J,\text{naive}}-\mathcal{F}^\star_{D,\text{naive}}|}$,
        we use cut decomposition to find suitable $D$ and apply Lemma~\ref{thm:naive_MF_bound}.
        
        \item \textbf{Step 2:}
        To bound \green{$|\ln \calZ_D - \mathcal{F}^\star_{D,\text{naive}}|$}, we introduce another $\gamma$-parametrized variational ansatz $\ln \calZ_{D,\gamma}^\star$ 
        such that
\begin{align}
    \green{
    \abs{\ln \calZ_D - \mathcal{F}^\star_{D,\text{naive}}}
    }
    \le 
    \abs{\ln \calZ_{D} - 
    \ln \calZ_{D,\gamma}^\star}
    +
    \abs{\ln \calZ_{D,\gamma}^\star-\mathcal{F}^\star_{D,\text{naive}}},\nonumber
\end{align}
        where $|\ln \calZ_{D}- 
        \ln \calZ_{D,\gamma}^\star|$ and $|\ln \calZ_{D,\gamma}^\star-\mathcal{F}^\star_{D,\text{naive}}|$ can be controlled  by using cut perturbation Lemma~\ref{lemma:E_decomposed_bound}.
        Rationale of introducing the ansatz $\ln \calZ_{D,\gamma}^\star$ is to approximate the intractable $\ln \calZ_D$ with its dominant term while other terms are exponentially suppressed.
        Physically, $\calZ_{D,\gamma}^\star$ can be understood as a tractable \textit{degenerated single-state} Ising model that serves as a perturbed approximation to $\calZ_{D}$ such that Lemma~\ref{lemma:E_decomposed_bound} is applicable to provide an analytic error bound.
    
        \item \textbf{Step 3:}
        Combining above and applying Lemma~\ref{lemma:trivial_bound_naiveMF}, we arrive the bound for $|\calF^\star_{J,\text{naive}}-\mathcal{F}|$ in the form of \eqref{ineq_3terms}.

        \item \textbf{Step 4:}
        From above, we complete the proof by obtaining $\calF_{J,\text{CoRMF}}^\star-\calF$ via Lemma~\ref{lemma:F_RNN}.

    \end{itemize}

\subsection{Step 1: Bounding $\abs{\ln \calZ_J -\ln \calZ_{D}}$ and $\abs{\mathcal{F}^\star_J-\mathcal{F}^\star_{D}}$}

We define the cut-decomposed matrix as  $D:=D^{(1)}+\dots+D^{(p)}$, where $D^{(1)},\cdots,D^{(p)}$ are the cut matrices coming from the Ising model $J$ according to Lemma~\ref{lemma:Frieze_Kannan} with the particular parameter choice $\epsilon\to\epsilon/12$ (i.e. $16/\epsilon^2\to 2304/\epsilon^2$), so that $p \le 2304/\epsilon^2$ and 
$$\norm{W}_{\infty\mapsto1}
=\|J - D\|_{\infty \mapsto 1} 
\leq \frac{\epsilon}{3} N\norm{J}_F.
$$ 
By Lemma~\ref{thm:naive_MF_bound}, it follows that 
\bea
\boxed{
\blue{\abs{\ln \calZ_J -\ln \calZ_{D}}\leq \frac{\epsilon}{3} N\norm{J}_F}, \quad \text{and} \quad
\Blue{\abs{\calF^\star_{J}-\mathcal{F}^\star_{D}}\leq \frac{\epsilon}{3} N\norm{J}_F }.
}
\label{eqn:bound1&2}
\eea

\subsection{Step 2: Bounding \green{$|\ln \calZ_D - \mathcal{F}^\star_{D,\text{naive}}|$} as $\abs{\ln \calZ_{D}-
\ln \calZ_{D,\gamma}^\star}$ and $\abs{\ln \calZ_{D,\gamma}^\star + \mathcal{F}^\star_D }$}

In order to bound \green{$|\ln \calZ_D - \mathcal{F}^\star_{D,\text{naive}}|$}, we introduce a analytically tractable variational term $\ln \calZ_{D,\gamma}^\star$ to divide it into two part as aforementioned, and use the cut perturbation Lemma~\ref{lemma:E_decomposed_bound} to control them one by one.
The intuition behind is, we introduce another auxiliary Ising model whose dominant term approximates $\calZ_D$ with an error that is analytically tractable, hence \textit{perturbation}.

In order to apply Lemma~\ref{lemma:E_decomposed_bound} and construct a bound for $\abs{\ln \calZ_{D}- 
    \ln \calZ_{D,\gamma}^\star}$, we do the follows:
\begin{claim} \vspace{-4mm}
\begin{itemize}
    \item For a given Ising model $J$, given a cut-decomposition $D$, we   map the configurations $\{\bX\}$ onto the $\vec{S}$-$\vec{T}$ plane, i.e. $\bX \to (\vec{S},\vec{T})$ where $(\vec{S},\vec{T})$ denotes realizations (points) on $\vec{S}$-$\vec{T}$ plane.
    
    \item
    For each point in $\vec{S}$-$\vec{T}$ plane, we define the corresponding local configuration set $\calX_{\Vec{S},\Vec{T},\gamma}$ in \eqref{eqn:local_set}.
    
    \item 
    We discretize $\vec{S}$-$\vec{T}$ with hyper-grid of size $\gamma N$,
    and use $\calX_{\Vec{S},\Vec{T},\gamma}$ to expand a hypercube of size $2\gamma N$ at each realization $(\vec{S},\vec{T})$.

    \item
    We approximate $\calZ_D^\star$ with its dominant term $\calZ_{D,\gamma}^\star$, where $\calZ_{D,\gamma}^\star$ is defined as a $\gamma$-parametrized variational anstaz capturing most contributing $(\vec{S},\vec{T})$ points. 

\end{itemize}
\end{claim}

We first explicitly write down the partition function of the \textit{cut-decomposed matrix} $D$
\bea
\label{eqn:Z_D_ST}
\calZ_{D}=
\sumX \exp{ -\sum_{\mu=1}^p \(\sum_{i\in [s]_\mu, j\in [t]_\mu} D_{ij}^{(\mu)} \bx_i\bx_j\)}
=
\sumX\exp{-\sum_{\mu=1}^{p}S_{\mu}(\bX)T_{\mu}(\bX)d_{\mu}},
\eea
with the shorthand notation $S_{\mu}(\bX)\coloneqq \sum_{i\in [s]_{\mu}}\bx_{i}\in \[-\abs{s_\mu},\abs{s_\mu}\]$ and $T_{\mu}(\bX)\coloneqq \sum_{j\in [t]_{\mu}}\bx_{j}\in \[-\abs{t_\mu},\abs{t_\mu}\]$.
By denoting $\Vec{S}(\bX)\coloneqq\(S_{1}(\bX),\cdots,S_{p}(\bX)\)$ the vector ranges over all elements of $\[-|s_{1}|,|s_{1}|\]\times\dots\times\[-|s_{p}|,|s_{p}|\]$ for each configuration $\bX$
(and similarly for $\Vec{T}(\bX)$), and replacing $\sumX$ with the sum over all possible $\vec{S}(\bX)$'s
and $\vec{T}(\bX)$'s realizations $\sum_{\Vec{S},\Vec{T}}$
, we obtain the inner-product representation for $\calZ_D$:
\bea
\label{eqn:Z_D_innerproduct}
\calZ_{D}
=\sum_{\Vec{S},\Vec{T}}\exp\left(-\sum_{\mu=1}^{p}S_{\mu}T_{\mu}d_{\mu}\right)\left(\sum_{\{\bX:\Vec{S}(\bX)=\Vec{S},\Vec{T}(\bX)=\Vec{T}\}}1\),
\eea
where $\sum_{\mu=1}^{p}S_{\mu}T_{\mu}d_{\mu}\coloneqq \bra{S_{\mu}}d_\mu\ket{T_{\mu}}$ denotes the $d_\mu$-weighted inner product between $\vec{S}$ and $\vec{T}$.
Here, in \eqref{eqn:Z_D_ST} and \eqref{eqn:Z_D_innerproduct}, we simply map the original intractable Ising model's configurations $\bX$'s onto the $\vec{S}$-$\vec{T}$ space where the partition function can be controlled analytically.

For any realization $\Vec{S} \in [-|s_1|,|s_1|]\times\dots\times[-|s_p|,|s_p|]$, $\Vec{T} \in [-|t_1|,|t_1|]\times\dots\times[-|t_p|,|t_p|]$ and the  $\gamma \in (0,1)$, 
we define the \textit{local configuration set} centered at $(\vec{S},\vec{T})$ with grid size $2\gamma N$ as 
\bea
\calX_{\Vec{S},\Vec{T},\gamma}
\coloneqq 
\big\{\bX\in \{\pm1\}^{N}: |S_\mu(\bX)-S_\mu| \leq \gamma N, |T_\mu(\bX)-T_\mu| \leq \gamma N \text{ for all } \mu \in [p]\big\},
\label{eqn:local_set}
\eea
such that 
$\(\vec{S}(\bX),\vec{T}(\bX)\)\big|_{\bX\in \calX_{\Vec{S},\Vec{T},\gamma}}$ 
denotes realizations close to $(\vec{S},\vec{T})$ within error $\gamma N$.

With $\calX_{\Vec{S},\Vec{T},\gamma}$, we further introduce the \textit{auxiliary hyper-grid} 
$I_\gamma \coloneqq \{\pm\gamma N, \pm 3\gamma N,\pm 5\gamma N,\cdots,\pm\ell \gamma N\}$ to fine-grain $\vec{S}$-$\vec{T}$ 
space with an auxiliary  $\vec{S}_\gamma $-$ \vec{T}_\gamma$ space, where $\ell$ is defined to be the smallest odd integer satisfying 
$\abs{\ell \gamma N - N} \leq \gamma N$
in order to completely capture all configurations in $\vec{S}$-$\vec{T}$ space, so that $\abs{\ell}\le 1/\gamma +1$ and $\abs{I_{\gamma}} \le 1/\gamma + 2$. 
In the auxiliary fine-grained $\vec{S}_\gamma $-$ \vec{T}_\gamma$ space, we define the following
the auxiliary Ising model
on $\vec{S}_{\gamma}$-$\vec{T}_\gamma$ as
\bea 
\label{eqn:True}
\calZ_{D,\gamma}
\coloneqq
\sum_{\vec{S}_\gamma,\vec{T}_\gamma\in I_{\gamma}^{p}}\[\exp{-\sum_{\mu=1}^{p}S_{\mu}^{(\gamma)}T_{\mu}^{(\gamma)}d_{\mu}}
\bigg(\abs{\calX_{\Vec{S}_\gamma,\Vec{T}_\gamma,\gamma}}\bigg)\]
,
\eea
and observe that the maximally contributed term in \eqref{eqn:True} is the following $\gamma$-parametrized lower bound of the optimal partition function of $D$, i.e. $\calZ_D^\star$ (the optimal value of $\calZ_D$ in $\vec{S}$-$\vec{T}$ space):
\bea
\label{eqn:Z_D_variational}
\calZ_{D,\gamma}^\star
\coloneqq
\max_{\vec{S}_\gamma,\vec{T}_\gamma\in I_{\gamma}^{p}}\exp\(-\sum_{\mu=1}^{p}S_{\mu}^{(\gamma)}T_{\mu}^{(\gamma)}d_{\mu}\)
\bigg(\abs{\calX_{\Vec{S}_\gamma,\Vec{T}_\gamma,\gamma}}\bigg)
,
\eea
where 
$\vec{S}_\gamma \coloneqq
(S_1^{(\gamma)},\cdots, S_p^{(\gamma)})$, 
$\vec{T}_\gamma \coloneqq (T_1^{(\gamma)},\cdots, T_p^{(\gamma)})$, $I_\gamma^p \coloneqq \overbrace{I_\gamma \times \cdots \times I_\gamma}^{p}$, and $|\calX_{\Vec{S}_\gamma,\Vec{T}_\gamma,\gamma}|$ 
counts the number of configurations ($\bX$'s or $(\vec{S},\vec{T})$'s, \textit{not} $(\vec{S}_\gamma,\vec{T}_\gamma)$'s) captured by the local configuration set centered at $(\vec{S}_\gamma,\vec{T}_\gamma)$, see 
Figure~\ref{fig:1}.
Intuitively, we should first note that, if we take the continuous limit $\gamma \to 0$, \eqref{eqn:Z_D_variational} recovers the term of optimal contribution from the RHS of \eqref{eqn:Z_D_innerproduct} and hence is a lower bound of $\calZ_D^\star$; that is,
\eqref{eqn:True} approximates the cut-Ising \eqref{eqn:Z_D_innerproduct} with controllable error characterizing by $\gamma$, and 
\eqref{eqn:Z_D_variational} can be regarded as the saddle approximation to $\calZ_{D,\gamma}$ by picking up the term with maximal contribution.

\begin{figure}[h]
  \centering
\includegraphics[width=0.5\textwidth]{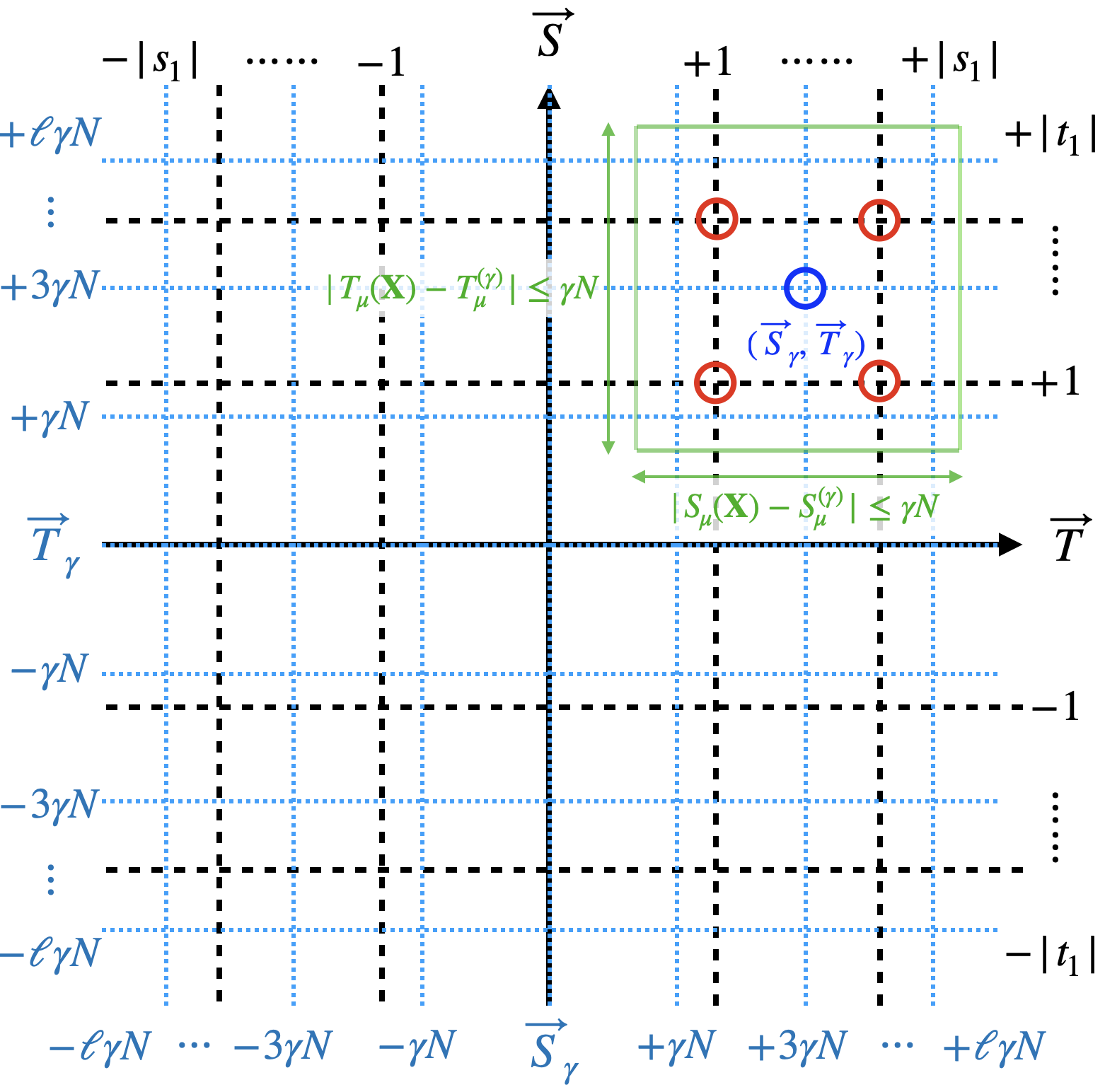}   
  \caption{\textbf{(Auxiliary Fine-Grained $\vec{S}_\gamma $-$ \vec{T}_\gamma$ Space.)}
  We consider the case of $p=1$ for simplicity.
  The black grid and blue grid represent the $\vec{S}$-$\vec{T}$ and $\vec{S}_\gamma$-$\vec{T}_\gamma$ spaces, respectively.
  Each intersection of black (blue) dashed-line is a realization of $\vec{S}$-$ \vec{T}$ ($\vec{S}_\gamma$-$\vec{T}_\gamma$) space.
  For any ($\vec{S}_\gamma$-$\vec{T}_\gamma$)-realization $(\vec{S}_\gamma,\vec{T}_\gamma)$ (the blue circle), we can expand a $\calX_{\vec{S}_\gamma,\vec{T}_\gamma,\gamma}$ (green box) around it capturing enclosed  $(\vec{S},\vec{T})$'s (red circles) with \eqref{eqn:local_set}, and compute $\calZ_{D,\gamma}^\star$ with \eqref{eqn:Z_D_variational}.
  }
  \label{fig:1}
\end{figure}

By Lemma~\ref{lemma:E_decomposed_bound}, we can construct $\abs{\ln \calZ_{D}-
\ln \calZ_{D,\gamma}^\star}$ by considering the following inequality
\begin{align}
\calZ^\star_{D,\gamma}e^{-8\norm{J}_F\gamma N\sqrt{p}} 
&\le 
\max_{\vec{S},\vec{T}}
\exp\left(-\sum_{\mu=1}^{p}S_{\mu}T_{\mu}
d_{\mu}\right)\left(\sum_{\{\bX:\Vec{S}(\bX)
=\Vec{S},\Vec{T}(\bX)=\Vec{T}\}}
1\)
\annot{(By Corollary~\ref{coro:exp_dst})}
\\
&\leq \calZ_{D} \label{eqn:Z_D_ineq1}
\\
&\leq
\sum_{\Vec{S}_\gamma,\Vec{T}_\gamma\in I_{\gamma}^{p}}\abs{\calX_{\Vec{S}_\gamma,\Vec{T}_\gamma,\gamma}} 
\exp{-\sum_{\mu=1}^{p}S_{\mu}^{(\gamma)}T_{\mu}^{(\gamma)} d_{\mu}}\exp{8\norm{J}_F\gamma N\sqrt{p}}  
\annot{(By Corollary~\ref{coro:dst_parity} \& \ref{coro:exp_dst})}
\\
& \leq   
\( \sum_{\Vec{S}_\gamma,\Vec{T}_\gamma\in I_{\gamma}^{p} } 1\)
\calZ^\star_{D,\gamma}
\exp{8\norm{J}_F\gamma  N\sqrt{p}}
= 
\abs{I_{\gamma}}^{2p}
\calZ^\star_{D,\gamma}  \exp{8\norm{J}_F\gamma N\sqrt{p}},\label{eqn:Z_D_ineq2}
\end{align}
where $\abs{I_{\gamma}}^{2p}$ is the total number of \textit{all} possible configurations in $\vec{S}_\gamma$-$\vec{T}_\gamma$ space\footnote{$\abs{I_{\gamma}}^{2p}$ is not necessarily the same as $|\{\bX\}|$ because we do not consider the 1-1 identification map from $\bX$ to $(\vec{S}(\bX),\vec{T}(\bX))$ here, i.e. the
$\left(\sum_{\{\bX|\Vec{S}(\bX)=\Vec{S},\Vec{T}(\bX)=\Vec{T}\}}1\)$ term in \eqref{eqn:Z_D_innerproduct}.}.
Consequently, from \eqref{eqn:Z_D_ineq1} and \eqref{eqn:Z_D_ineq2} we get
\begin{equation}
\label{eqn:Z_D_upper}
\ln \calZ^\star_{D,\gamma}\leq\ln \calZ_{D}+8\norm{J}_F\gamma N\sqrt{p},
\end{equation}
and 
\begin{align}
\label{eqn:Z_D_lower}
\ln \calZ_{D,\gamma}^\star
&\geq\ln \calZ_{D}-8\norm{J}_F\gamma N\sqrt{p} - 2p\ln \abs{I_\gamma} \nonumber\\
&\ge \ln \calZ_{D}-8\norm{J}_F\gamma N\sqrt{p} - 2p\ln\(\frac{1}{\gamma} + 2\),
\end{align}
respectively.
These bounds allow us to control $\abs{\ln \calZ_{D}-\ln \calZ_{D,\gamma}^\star}$ by
\bea
\green{
\boxed{
\abs{\ln \calZ_{D}-\ln \calZ_{D,\gamma}^\star}
\le
8\norm{J}_F\gamma N\sqrt{p} + 2p\ln\(\frac{1}{\gamma} + 2\),
}
}
\label{eqn:bound3}
\eea
with variational parameter $\gamma\in(0,1)$ whose optimal value will be determined shortly in \textbf{Step 3}.

In order to construct bounds to control $\abs{\mathcal{F}^\star_D -\ln \calZ_{D,\gamma}^\star}$,
we need to develop mean-field description for $\ln \calZ_{D,\gamma}^\star$.
That is, while \eqref{eqn:Z_D_variational} only has one configuration in $\vec{S}_\gamma$-$\vec{T}_\gamma$ space, it corresponds to multiple $\bX$'s captured by $\calX_{\Vec{S}_\gamma,\Vec{T}_\gamma,\gamma}$ (the green box in Figure~\ref{fig:1}).
Naturally, for the optimal realization $(\vec{S}_\gamma,\vec{T}_\gamma)$,  \eqref{eqn:Z_D_variational} can be further interpreted as the inner-product representation of a \textit{local} Ising model restricted by $\calX_{\Vec{S}_\gamma,\Vec{T}_\gamma,\gamma}$. 
It is a \textit{degenerated one-state Ising model}.
By degenerated one-state, we mean this Ising model has only one energy state (only one $(\vec{S}_\gamma,\vec{T}_\gamma)$ realization), while there can be multiple configurations in it (multiple $\bX$.)

We define the corresponding auxiliary (naive) mean-field value, following \eqref{eqn:naive_MF},
\bea
\bar{y}_j \coloneqq \frac{1}{\abs{\calX_{\Vec{S}_\gamma,\Vec{T}_\gamma,\gamma}}} \sum_{\bX \in \calX_{\Vec{S}_\gamma,\Vec{T}_\gamma,\gamma}} \bx_j,
\quad \(\;\bar{Y}\coloneqq \( \bar{y}_1,\cdots,\bar{y}_N\)\;\)
\eea
as the mean value of configurations captured by $\calX_{\Vec{S}_\gamma,\Vec{T}_\gamma,\gamma}$.
Moreover, we denote probabilities associated to the \textit{restricted} configurations in $\calX_{\Vec{S}_\gamma,\Vec{T}_\gamma,\gamma}$ for such a \textit{local} and \textit{restricted} Ising model as
$\bY \coloneqq (\by_1, \cdots, \by_N)\in [0,1]^N$.
As a result, 
the term 
$\abs{\calX_{\Vec{S}_\gamma,\Vec{T}_\gamma,\gamma}}$ 
is nothing but the entropy $H(\bY)$ of the \textit{local} and \textit{restricted}  auxiliary Ising model $\calZ_D^\star$.
Therefore, we have
\begin{align}
\ln \abs{\calX_{\Vec{S}_\gamma,\Vec{T}_\gamma,\gamma}} &= H(\bY) \label{eqn:H_Y}
\\
&\le \sum_{j = 1}^N H(\by_j) 
= \sum_{j = 1}^N H\(\frac{1 + \overline{y}_j}{2}\),
\annot{(By \eqref{eqn:naive_MF})}
\end{align}
where the independent decomposition property of naive mean-field is used in the second line.

Taking logarithm of \eqref{eqn:Z_D_variational}, we have 
\begin{align}
\label{eqn:bound3.5}
\ln \calZ_{D,\gamma}^\star 
& =
-\sum_{\mu=1}^{p}S_{\mu}^{(\gamma)}T_{\mu}^{(\gamma)} d_{\mu}
+\ln\abs{\calX_{\Vec{S}_\gamma,\Vec{T}_\gamma,\gamma}}
\\
& \leq
-\sum_{\mu=1}^{p} S_{\mu}^{(\gamma)}T_{\mu}^{(\gamma)} d_{\mu}
+\sum_{j = 1}^N H\left(\frac{1 + \overline{y}_j}{2}\right)
\annot{(By \eqref{eqn:H_Y})}
\\
 & \leq
 \[-\sum_{\mu=1}^{p} S_{\mu}^{(\gamma)}T_{\mu}^{(\gamma)} d_{\mu}+8\norm{J}_F\gamma N \sqrt{p}
 \]
 +\sum_{j = 1}^N H\left(\frac{1 + \overline{y}_j}{2}\right)
 \annot{(By Lemma~\ref{lemma:E_decomposed_bound})}
\\
 & \leq
 \[ 
 \sum_{\mu=1}^{p}
 -S_{\mu}\(\bar{Y}\)T_{\mu}(\bar{Y})d_{\mu}+8\norm{J}_F\gamma N \sqrt{p}\] + \sum_{j = 1}^N H\left(\frac{1 + \overline{y}_j}{2}\right)
 \annot{(By Jensen's inequality)}
 \\
 & \leq -\mathcal{F}_{D}^\star+8\norm{J}_F\gamma N\sqrt{p}.
 \annot{(By \eqref{eqn:naive_MF})}
\end{align}
By combining \eqref{eqn:bound3} and \eqref{eqn:bound3.5}, we are able to control $\green{
\abs{\ln \calZ_{D}- 
    \ln \calZ_{D,\gamma}^\star}
+\abs{\ln \calZ_{D,\gamma}^\star+\mathcal{F}^\star_D}
}$.
Firstly, one notice that
\begin{align} 
-\mathcal{F}_{D}^\star 
& \geq \ln \calZ_{D,\gamma}^\star - 8\norm{J}_F\gamma N\sqrt{p}\nonumber
\\
& \geq \ln \calZ_{D} - 16\norm{J}_F\gamma N\sqrt{p} - 2p\ln(\frac{1}{\gamma} + 2)
\annot{(By \eqref{eqn:Z_D_lower})}
\\
 & \geq \ln \calZ_{D} - \frac{\epsilon }{3}N\norm{J}_F - 9216\ln\left(\frac{48}{\epsilon} + \sqrt{2}\right)\frac{1}{\epsilon^{2}},
\end{align}
where in the last line,
for any $\epsilon > 0$,
 we choose $\gamma=\epsilon/(48\sqrt{p})$ and $p\le 2304/\epsilon^2$ in order to obtain the $\epsilon/3$ factor in front of the second term, and the inequality $(a+b)^2\le a^2+b^2$.
Secondly, by noticing $\mathcal{F}^\star_D + \ln \calZ_{D}\geq 0$,
we arrive 
\bea
0 \leq \ln \calZ_{D}+\mathcal{F}^\star_D \leq   \frac{\epsilon }{3}N\norm{J}_F + 9216\ln\left(\frac{48}{\epsilon} + \sqrt{2}\right)\frac{1}{\epsilon^{2}},
\eea
which implies
\bea
\label{eqn:bound4}
\green{
\boxed{
\abs{\ln \calZ_{D}- 
    \ln \calZ_{D,\gamma}^\star}
+\abs{\ln \calZ_{D,\gamma}^\star+\mathcal{F}^\star_D}
\le 
\green{
\frac{\epsilon}{3}\epsilon n\norm{J}_F + 9216\ln\left(\frac{48}{\epsilon} + \sqrt{2}\right)\frac{1}{\epsilon^{2}}.
}
}}
\eea

\subsection{Step 3: Bounding $\abs{\calF^\star_{J,\text{naive}}-\mathcal{F}}$}
Combining above bounds \eqref{eqn:bound1&2}, \eqref{eqn:bound3} and \eqref{eqn:bound4}, for any $\epsilon > 0$, we get
\begin{align}
\calF^\star_{\text{naive}}-\mathcal{F}
&\leq  
\blue{
\abs{\ln \calZ -\ln\calZ_{D}}
}
+
\Blue{
\abs{\mathcal{F}^\star-\mathcal{F}^\star_{D}}
}
+
\green{
\abs{\ln \calZ_{D}- 
    \ln \calZ_{D,\gamma}^\star}
+\abs{\ln \calZ_{D,\gamma}^\star+\mathcal{F}^\star_D}
}
    \nonumber\\
& \le 
\blue{
\frac{\epsilon}{3} N \norm{J}_F
}+
\Blue{
\frac{\epsilon}{3} N \norm{J}_F 
}
+
\green{
\frac{\epsilon}{3} N \norm{J}_F + 9216\ln\left(\frac{48}{\epsilon} + \sqrt{2}\right)\frac{1}{\epsilon^{2}}
}
\nonumber\\
& \leq \epsilon N \norm{J}_F  + 9216\ln\left(\frac{48}{\epsilon} + \sqrt{2}\right)\frac{1}{\epsilon^{2}}<
\epsilon N \norm{J}_F  + 9216\ln\left(\frac{48}{\epsilon}+e\right)\frac{1}{\epsilon^{2}},
\label{eqn:bound5}
\end{align}
where we introduce $e$ in the last line to ensure $\ln(\cdot)\ge 1$. 

Before getting technical, it is convenient to introduce some intuitions about how to pick $\epsilon$ in order to minimize \eqref{eqn:bound5}:
If we observe carefully, 
it is clear that $\epsilon$ is a function of $N$ and $\norm{J}_F$, and  
the LHS of \eqref{eqn:bound5} asymptotically scales with $\calO\(\(N\norm{J}_F\)^{2/3}\)$.

To see this, we consider the following problem of minimizing the RHS of \eqref{eqn:bound5}. 
Naturally, we let two competing terms (two terms with opposite scalings with $\epsilon$) on the RHS of \eqref{eqn:bound5} to be the same. 
However, we can not solve $\epsilon N \norm{J}_F=9216\ln\left(48/\epsilon+e\right)1/\epsilon^{2}$ directly. 
Instead, by simply verifying ratio of first order derivatives\footnote{$\lim_{\epsilon \to 0 } \nicefrac{\frac{\dd \(1/\epsilon^2\) }{\dd \epsilon }}{\frac{\dd \ln\(1/\epsilon\) }{\dd \epsilon }} =\lim_{\epsilon \to 0 }\nicefrac{2}{\epsilon^2}= \infty.$}, we notice that absolute value of the slope of  $\ln\(1/\epsilon\)$ is smaller than absolute value of the slope of  $1/\epsilon^2$, and hence the latter dominates the bound.
By neglecting the $\ln$ term and letting two competing terms be the same, we have $\epsilon \propto \(N \norm{J}_F\) ^{-1/3}$. 
Plugging $\epsilon$ back to \eqref{eqn:bound5}, 
we find the bound is asymptotically proportional to  $(N\norm{J}_F)^{2/3}$ modulo $\ln$ term. 
Therefore, to obtain the explicity functional form of $\epsilon\(N\norm{J}_F\)$, we need to compare $1/\epsilon$ with $N\norm{J}_F$.
In the following, we examine above intuition in detail  for all $N\norm{J}_F.$

Our immediate goal is to choose the optimal $\epsilon$ that gives the tightest bound for  \eqref{eqn:bound5}, where the two terms compete with each other. 
We discuss the following two cases:
\begin{itemize}
    \item 
    First we consider 
    \bea \label{condi-1}
    \frac{1}{\epsilon}<N \norm{J}_F.
    \eea
    Since $\gamma\in (0,1)$, $\gamma=\epsilon/(48\sqrt{p})$, and $p\le 2304/\epsilon^2$, we have
    \bea
    0<\epsilon< \Min\left\{48 \sqrt{p}, \sqrt{\frac{2304}{p}}\right\},
    \eea
    indicating $\epsilon$ is upper bounded and hence 
    $\epsilon \propto \(N \norm{J}_F\) ^{-1/3}$ implies $N\norm{J}_F$ lower bounded.
    We let $M \geq e$ a fixed lower bound for $N\norm{J}_F$ such that $N\norm{J}_F>M.$
    
    Motivated by \cite{jain2018mean}, we take $\epsilon = \left(\nicefrac{M\ln(48 N \norm{J}_F + e)}{N \norm{J}_F \ln{M}}\right)^{1/3},$
    which leads to the following inequality
    \begin{align}
        \label{tri}
    \frac{\ln M}{M} \leq 1 \leq N^2\norm{J}_F^2 \ln\(48 N\norm{J}_F+e\),
    \end{align}
    By $M>1$ and $\ln(48 N\norm{J}_F+e)>1$.
Thus we can rewrite \eqref{condi-1} as
\bea
\frac{1}{\epsilon}
=N^{1/3}\norm{J}_F^{1/3} \(\frac{\ln M}{M \ln(48 N\norm{J}_F+e)}\)^{\nicefrac{1}{3}} \leq N \norm{J}_F.
\eea

We also have
\begin{align}
    \ln\left(\frac{48}{\epsilon}+e\right) \leq  \ln\left(48 N \norm{J}_F+e\right).
    \annot{(By \eqref{condi-1})}
\end{align}
So \eqref{eqn:bound5} becomes
\bea
\calF^\star_{\text{naive}}-\mathcal{F} 
&\leq& \(\frac{M}{\ln M}\)^{\nicefrac{1}{3}} N^{\nicefrac{2}{3}} \norm{J}_F^{\nicefrac{2}{3}} \ln^{\nicefrac{1}{3}}\(48N \norm{J}_F + e\)\\  &&+ 9216\(\frac{\ln M}{M}\)^{\nicefrac{2}{3}} \frac{\ln(48 N \norm{J}_F + e)}{\ln^{\nicefrac{2}{3}}\(48 N \norm{J}_F + e\)} N^{\nicefrac{2}{3}} \norm{J}_F^{\nicefrac{2}{3}} \nonumber \\
&\leq&
N^{\nicefrac{2}{3}} \norm{J}_F^{\nicefrac{2}{3}} \ln^{\nicefrac{1}{3}}\(48 N \norm{J}_F + e\)
\[\(\frac{M}{\ln M}\)^{\nicefrac{1}{3}} + 9216 \(\frac{\ln M}{M}\)^{\nicefrac{2}{3}}\].\nonumber
\eea
Finally, by taking $\(M/\ln M\) = 9216$ without loss of generality, we find
\bea
\mathcal{F}^\s_{\text{naive}}- \mathcal{F}
\le 42 N^{\nicefrac{2}{3}} \norm{J}_F^{\nicefrac{2}{3}} \ln^{\nicefrac{1}{3}}(48 N \norm{J}_F + e)\propto \(N \norm{J}_F\)^{\nicefrac{2}{3}}. 
\eea
    \item
    Now we can consider another case 
    \bea
    N\norm{J}_F \leq M.
    \eea
    By Lemma~\ref{lemma:trivial_bound_naiveMF},
    we have $\calF^\star_{\text{naive}}-\mathcal{F} \leq N\norm{J}_F$. 
    Therefore, if $N \norm{J}_F \leq M$, we arrive
    \bea
    \calF^\star_{\text{naive}}-\mathcal{F} \le N \norm{J}_F \leq M^{\nicefrac{1}{3}} \(N \norm{J}_F\)^{\nicefrac{2}{3}} \propto \(N \norm{J}_F\)^{\nicefrac{2}{3}}.
    \eea
\end{itemize}

\subsection{Step 4: Bounding $\calF_{J,\text{CoRMF}}^\star-\calF$ with Lemma~\ref{lemma:F_RNN}}
We then combine two cases and show that for all $N \norm{J}_F$ and then restore all coefficients. 
So we have
\bea
\calF^\star_{\text{CoRMF}}-\mathcal{F}\leq 
\calF^\star_{\text{naive}}-\mathcal{F} 
\le 42 N^{\nicefrac{2}{3}} \norm{J}_F^{\nicefrac{2}{3}} \ln^{\nicefrac{1}{3}}\(48 N \norm{J}_F + e\),
\eea
or with $\beta$
\bea
\calF^\star_{\text{CoRMF}}-\mathcal{F}\leq 
\calF^\star_{\text{naive}}-\mathcal{F} \leq \frac{1}{\beta} \calO\(N^{\nicefrac{2}{3}} \norm{\beta \tilde{ J}}_F^{\nicefrac{2}{3}} \ln^{\nicefrac{1}{3}}\(N \norm{\beta \tilde{J}}_F\)\),
\eea
which completes the proof. 
\end{proof}

\section{Numerical Experiments Details}
\label{sec:exp_details}

\subsection{CoRMF Details}
\begin{itemize}
    \item 
    \textbf{Architectural Details:}
    We implement CoRMF in an atomic setting a with 2-layer RNN (with $\sim 5000$ parameters) as the Recurrent Network Module for N=10,20 CoRMF experiments in this work.
    This choice was made to avoid adding extra inductive bias and assumptions from the architecture design, as our only assumption regarding the architecture is the recurrent parametrization.
    We emphasis that the choice of RNN module is fully customizable.
    The number of features in the hidden state of RNN is set to 50 for all experiments. A fully connected layer with a Softmax  activation layer is appended after the RNN to process the hidden feature and output the conditional probabilities.

    \item 
    \textbf{Training Details:}
        \begin{itemize}
            \item 
            \textbf{Optimizer:}
            We use an Adam optimizer \cite{kingma2015adam} with learning rate $lr=0.001$ for training. 
            The coefficients of Adam optimizer, betas, are set to $(0.9,0.999)$. 
            We also use a scheduler to adjust the learning rate of this optimizer. 
            It reduces learning rate when the free energy $\calF^\star$ stopped improving in training process. 
            The patience of this scheduler is set to 1000 and the decay factor is set to 0.8.
            We set the batch size to $1000$, and the total number of iterations to $10000$.
            We use the same optimizer for the NMF baseline.
            \item 
            \textbf{$\beta$ Annealing:}
            Following \cite{nicoli2020asymptotically}, we use $\beta$-annealing technique to avoid mode trapping at early stage of training.
            \item
            \textbf{Gradient Norm Clipper:} We clips gradient norm of our model's parameters to resolve potential gradient explosion problem. 
            The  maximum allowed value of the gradients is set to 1.

            \item
            \textbf{Gradient Estimator:}
            We set the sample size $K$ of gradient estimator to $1000$.
            
        \end{itemize}
        \item 
        \textbf{Platforms:}
        The GPUs and CPUs used to perform CoRMFs and evaluate the latency of baseline are NVIDIA GEFORCE RTX 2080 Ti and INTEL XEON SILVER 4214 @ 2.20GHz.
\end{itemize}

\subsection{Datasets}
For Ising models we use in experiments, we scale $(J,h)$ with some of them to avoid the system being ``too deterministic" (only one configuration appears while all others are suppressed exponentially.)
We prefer some randomness to make the 
Meanwhile, we still perform the N=100 Spin Chain as a deterministic example. 
\begin{itemize}
    \item 
    \textbf{N=100 Ising Spin Chain:}
    We set $J_{ij}=-1$ for all neighboring $(i,j)$.
    We set $h_i=1$ for all $i$ to enforce $\Braket{\bx}_Q$ deviate from $0$ such that evaluation becomes easier.

    \item 
    \textbf{N=10 Ising Model:}
    \begin{itemize}
        \item 
        \textbf{$\beta=1$:}
        $(J_{ij},h_i)$ are picked from $[55]/10$ without replacement, and $h_i$'s are strengthen by a factor of $1.3$ in order to make $\Braket{\bx}_Q$ evaluation easier.
        \item
        \textbf{$\beta=5$:}
        Recaling $(J,h)$ of above $\beta=1$ Ising model with a factor of $5$.

    \end{itemize}

    \item 
    \textbf{N=20 Ising Models (without external fields):}
    For N=20 Ising models, we turn off the external field by setting $h=0$ so that outcomes are all caused by spin-spin interactions, and the effects of criticality order are isolated.
    \begin{itemize}
        \item 
        \textbf{Dense N=20 with L=400:}
        We set $J_{ij}\sim \calU([L=N^2=400])/100$ with $\calU([L])$ denoting ``uniformly pick an integer between $1$ to $L$."
        In this setting, the criticality order is rarely contaminated and the probability of picking repeated number is considerably low as $L\ge N^2$. 

        \item
        \textbf{Dense N=20 with L=5:}
        We set $J_{ij}\sim \calU([L=5\ll N^2])/2$
        According to the preceding definition, the $L=5$ setting causes each number to appear $80$ times on average.
        We regard the order obtained in this setting is \textit{highly ambiguous} due to large number of repeated $J_{ij}$.

        \item
        \textbf{Sparse N=20:}
        To introduce sparsity into the Ising graph, we simply set $J_{ij}\sim \mathsf{Poisson}(\lambda=0.4)$ with $P(J_{ij}=0)\simeq 0.67$.

        \item
        \textbf{Random N=20:}
        To introduce more complicated, sparse and ambiguous interaction pattern, we  set $J_{ij}\sim \calU([L=5]/2-1)$.
    \end{itemize}

\end{itemize}

\subsection{Reference Values}
The reference values of $\calF^\star$ and $\Braket{\bx}_Q$ are obtained by sampling $P(\bX)\propto e^{-\beta E(\bX)}$ with MCMC Gibbs sampler.
After checking the number burn-in iterations of all datasets, 
for each set of $(J,h)$, we repeatedly do (i) run the number of burn-in iterations and (2) sample \textit{one} configuration.
For all Ising models, we use 10,000 samples to compute reference values of our interest.

\subsection{Visualizations}
To visualize the training process and the quality of proposed method, we plot the variational free energy evolution and mean parameter of spins (average over sampled configurations)  over iterations for the \textit{N=100 1D Spin Chain}, \textit{N=10 Ising ($\beta=1$)} and \textit{Dense N=20 Ising (L=400)} experiments as follows.

\begin{figure}[h]
  \centering
  {\includegraphics[width=0.3\textwidth]{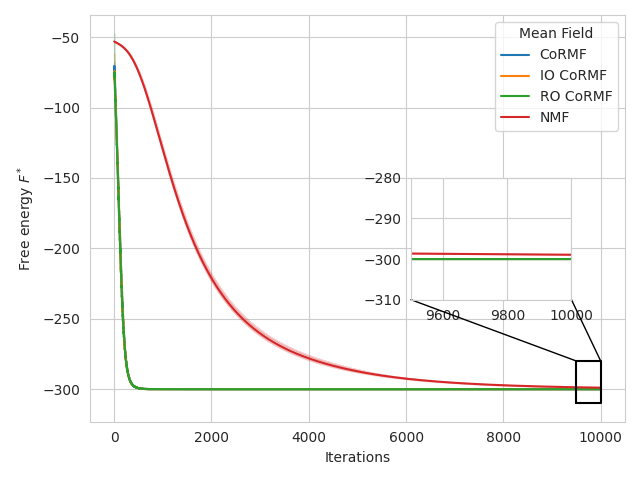}
  \label{fig:SpinChain}}
  \hfill
  {\includegraphics[width=0.3\textwidth]{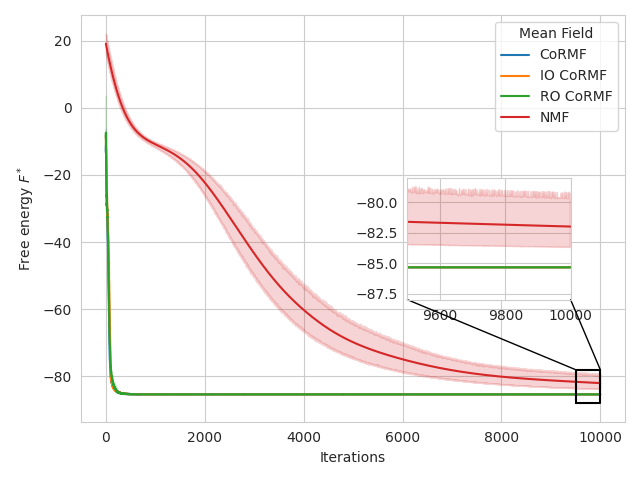}
  \label{fig:N10beta1}}
  \hfill
  {\includegraphics[width=0.3\textwidth]{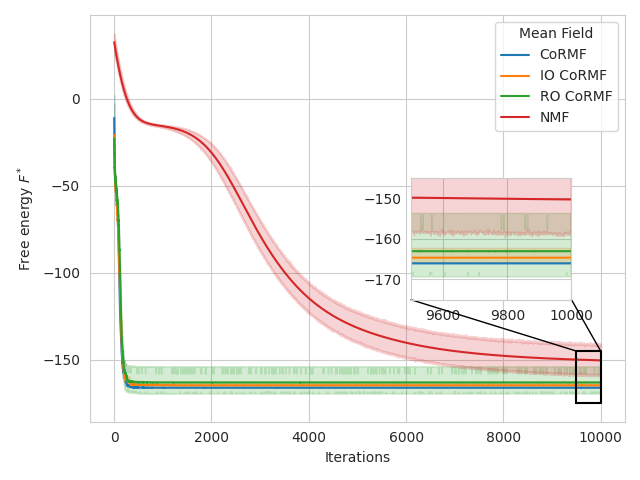}
  \label{fig:denseN20_L400}}
  \\ \vspace{-0.2Truein}
  \subfloat[N=100 1D Spin Chain.]{\includegraphics[width=0.3\textwidth]{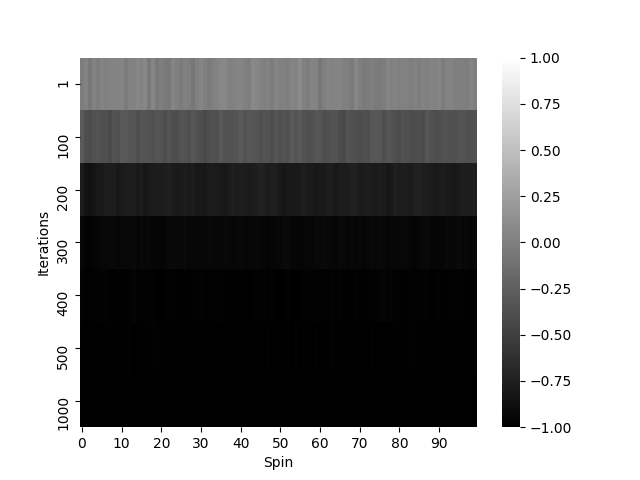}
  \label{fig:ll}}
  \hfill
  \subfloat[N=10 Ising ($\beta$=1).]{\includegraphics[width=0.3\textwidth]{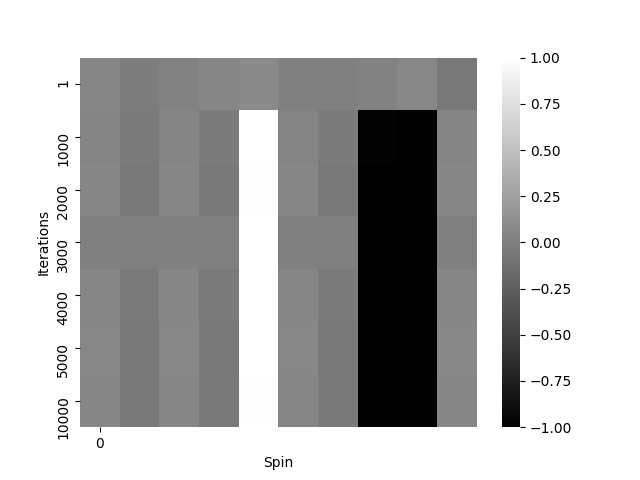}
  \label{fig:mag_N10scale}} 
  \hfill
  \subfloat[Dense N=20 Ising (L=400).]{\includegraphics[width=0.3\textwidth]{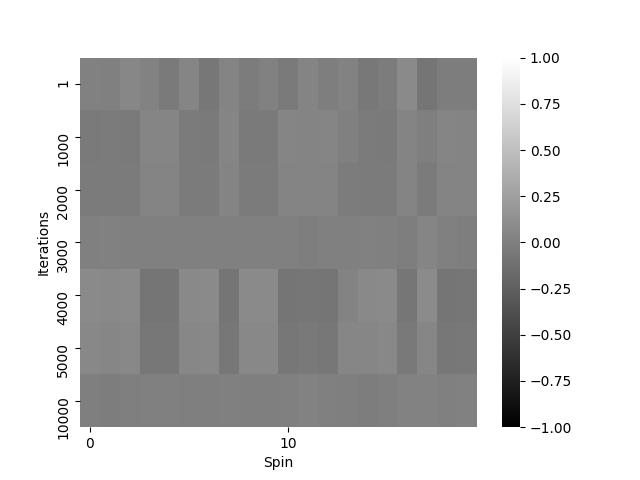}
  \label{fig:44}}  
  \caption{\textbf{(Variational Free Energy $\calF^\star$ \& Mean Parameter of Spins $\E_Q[\bx_i]$)}
  In the upper row of the plots, the $\calF^\star$ evolution curves, we can easily tell that the CoRMF with different orders (criticality-, inverse-, and random-) outperform the NMF baseline by faster and lower convergence (thus being more efficient, accurate, and aligned with our theoretical finding, Theorem~\ref{thm:main_thm}).
  In the lower row, the $\E_Q[\bx_i]$  over iterations for all $i$ are provided where $Q$ is governed by CoRMF.
  We observe that, for each spin, $\E_Q[\bx_i]$ becomes stabilized once the convergence is attained in both $h\neq 0$, (a) \& (b), and $h=0$, (c), cases.
  In (a), we see that the trivial interaction (nearest-neighbor) in 1D results to almost deterministic spin patterns ($\E_Q[\bx_i]\simeq-1$ for all spins) within 300 iterations.
  In (b) and (c), we see that more complicated interactions presents richer spin patterns, and the significance of the tree-structured order becomes noticeable (the lines split from each other as we zoom in.)
  }
\end{figure}

\clearpage
\section*{Related Works}

\paragraph{Deep Learning Enhanced Forward Ising Solver.}
The exact inference of many discrete models is NP-hard \cite{wainwright2008graphical,koller2009probabilistic}, including Ising model \cite{sly2012computational,istrail2000statistical,barahona1982computational}.
To handle this, our works is motivated by and built on the series works of variational inference on continuous \cite{kingma2013auto,blei2017variational,mnih2016variational,ranganath2014black} and discrete \cite{bengio2013estimating,rolfe2016discrete}  domains that involve approximating posteriors with neural networks. 
More specifically, our work falls under the \textit{Variational Mean Field} (VMF) category \cite{attias1999variational,wainwright2008graphical}, which has its roots in statistical physics \cite{weiss1907hypothese,kadanoff2009more}, is one of the simplest and most prominent variational approximations to the free energy (the log partition function) $\ln \calZ$.
Searching for richer variational families beyond the fully factored Naive Mean-Field (NMF) would enable a tighter bound on the free energy, and hence limit the likelihood of evidence \cite{wiegerinck2013variational,xing2012generalized,yedidia2003understanding,jordan1999introduction}.
In this work, the problem setting of our interest, the forward Ising, does not involve any conditional evidence. 
Therefore, to our best knowledge, our method, which is a data-free \& neutralized variant of VMF, presents and solves an unique type of problem that is
less discussed in variational inference literature.
We view CoRMF as an attempt to develop a forward Ising solver powered by deep learning.

\paragraph{Forward Ising Problems and Ising Machines.}
We discuss the existing hardware Ising solvers here.
With Ising formulation, or equivalently Quadratic Unconstrained Binary Optimization (QUBO) \cite{kochenberger2006unified}, for many NP problems \cite{lucas2014ising}, many systems have been proposed to solve NP problems as forward Ising problems, including D-Wave’s quantum annealers \cite{king2021scaling,ushijima2017graph,bian2016mapping,bian2014discrete}, Coherent Ising Machines \cite{inagaki2016coherent}, Electronic Oscillator-based Ising Machines (OIM) \cite{vaidya2022creating,wang2019new}, and BRIM \cite{afoakwa2021brim}, an CMOS-compatible Ising Machine. 
A work \cite{reneau2023feature,zhao2018stock} also investigate the possibility of using Ising model to forecast stock prices (and time series) and other complicated problems. While these studies theoretically demonstrate the potential of mapping such applications onto the Ising models, in practice, they fail to achieve similar accuracy to conventional approaches, and computational performance metrics like latency \cite{son2006random} and throughput are seldom reported. And with the support of increasingly powerful Ising machines \cite{afoakwa2021brim}, recent work \cite{pan2023ising, liu2023ising} firstly propose Ising approaches outperforming the Neural Networks in both accuracy and real-time perspectives. They also demonstrate that it is the time to develop comprehensive domain-specific Ising models for the real world.
Though significant progress has been made towards solving hard forward problems (e.g., on classical and quantum hardware solvers), the main challenge is still to handle hard problems within an affordable timeframe (and computational budge).
On one hand, the classical annealers fail to fulfill as it requires exponential time to reach the true global solution \cite{bertsimas1993simulated,van1987simulated}. 
On the other hand, we are still in an early stage of full-fledged quantum machine (beyond \textit{NISQ} \cite{preskill2018quantum,hu2022design}) that can solve hard problems within polynomial time without \textit{noise} \cite{cerezo2022challenges,bittel2021training,preskill2021quantum}.
To this end, our work serves as an on-Turing-machine alternatives to these classical and quantum Ising machines.

\paragraph{From Ising Energy Minimization to Deep Learning and Foundation Model.}
Recently, the intersection of energy minimization in the Ising model with brain science has propelled deep learning forward, especially in foundational model research, as seen through the Hopfield model \cite{hopfield1982neural,hopfield1984neurons}. This model, by leveraging energy minimization, effectively mimics neural processes in the brain for storing and retrieving memory patterns. Its adaptation into deep learning brings associative memory and energy minimization concepts to the forefront, offering a nuanced interpretation of transformer-attention neural architectures \cite{hu2024computational,wu2023stanhop,hu2023SparseHopfield,ramsauer2020hopfield}. This approach enriches our comprehension of transformers' core components \cite{vaswani2017attention}, paving the way for innovative Hopfield-inspired architectural designs. 
Consequently, these insights extend their utility across various domains, including drug discovery \cite{schimunek2023contextenriched}, immunology \cite{widrich2020modern}, time series forecasting \cite{wu2023stanhop,auer2023conformal}, reinforcement learning \cite{paischer2022history}, and large language models \cite{furst2022cloob}.

\clearpage
\bibliography{main_aistats.bib}
\bibliographystyle{unsrtnat}

\end{document}